\documentclass[journal]{IEEEtran}
\usepackage{xcolor}
\usepackage{amsmath}
\usepackage{amssymb}
\usepackage{multirow}
\usepackage{booktabs}
\usepackage{makecell}
\newcommand{\ra}[1]{\renewcommand{\arraystretch}{#1}}
\usepackage{graphicx}
\usepackage{arydshln}

\newcommand{\mypm}{\mathbin{\smash{%
\raisebox{0.35ex}{%
            $\underset{\raisebox{0.5ex}{$\smash -$}}{\smash+}$%
            }%
        }%
    }%
}

\ifCLASSINFOpdf
\else
\fi
\begin{document}
\title{Age-Oriented Face Synthesis with Conditional Discriminator Pool and Adversarial Triplet Loss}

\author{Haoyi~Wang,
        Victor~Sanchez,~\IEEEmembership{Member,~IEEE,}
        Chang-Tsun~Li,~\IEEEmembership{Senior~Member,~IEEE}% <-this % stops a space
\thanks{H. Wang and V. Sanchez are with the Department
of Computer Science, University of Warwick, Coventry, CV4 7AL, UK (e-mail: h.wang.16@warwick.ac.uk, v.f.sanchez-silva@warwick.ac.uk.)}% <-this % stops a space
\thanks{C-T. Li is with the School of Information Technology, Deakin University, Geelong VIC 3216, Australia (e-mail: changtsun.li@deakin.edu.au.)}% <-this % stops a space
\thanks{© 2020 IEEE.  Personal use of this material is permitted.  Permission from IEEE must be obtained for all other uses, in any current or future media, including reprinting/republishing this material for advertising or promotional purposes, creating new collective works, for resale or redistribution to servers or lists, or reuse of any copyrighted component of this work in other works.}% <-this % stops a space
% \thanks{Manuscript received April 19, 2005; revised August 26, 2015.}}
}


\markboth{Journal of \LaTeX\ Class Files,~Vol.~14, No.~8, August~2015}%
{Shell \MakeLowercase{\textit{et al.}}: Bare Demo of IEEEtran.cls for IEEE Journals}



\maketitle

\begin{abstract}
The vanilla Generative Adversarial Networks (GAN) are commonly used to generate realistic images depicting aged and rejuvenated faces. However, the performance of such vanilla GANs in the age-oriented face synthesis task is often compromised by the mode collapse issue, which may result in the generation of faces with minimal variations and a poor synthesis accuracy. In addition, recent age-oriented face synthesis methods use the L1 or L2 constraint to preserve the identity information on synthesized faces, which implicitly limits the identity permanence capabilities when these constraints are associated with a trivial weighting factor. In this paper, we propose a method for the age-oriented face synthesis task that achieves a high synthesis accuracy with strong identity permanence capabilities. Specifically, to achieve a high synthesis accuracy, our method tackles the mode collapse issue with a novel Conditional Discriminator Pool (CDP), which consists of multiple discriminators, each targeting one particular age category. To achieve strong identity permanence capabilities, our method uses a novel Adversarial Triplet loss. This loss, which is based on the Triplet loss, adds a ranking operation to further pull the positive embedding towards the anchor embedding resulting in significantly reduced intra-class variances in the feature space. Through extensive experiments, we show that our proposed method outperforms state-of-the-art methods in terms of synthesis accuracy and identity permanence capabilities, qualitatively and quantitatively.

% Although faces are composed of numerous facial attributes, most works with CNNs only consider faces as a  and do not pay enough attention to those regions that carry age-specific features for this particular task. In this paper, we propose a novel CNN architecture called Cumulative Convolutional Neural Network (Cumulative-CNN) to tackle the age estimation problem. Apart from the whole face image, the Cumulative-CNN successively takes several adaptively selected age-specific facial patches as part of the input to emphasize the age-specific features. Features learned from different sources are merged into the main information flow of the network, and the predicted age is obtained based on those accumulated features. Furthermore, through experiments, we show that the Cumulative-CNN significantly outperforms other state-of-the-art models on the MORPH II benchmark.
\end{abstract}


% \fi
\IEEEpeerreviewmaketitle

\section{Introduction}

% outline
% 1. what the face-based age estimation is about
% 2. people start to use patch based methods, like ...
% 3. drawback of the above methods, i.e. they use fixed facial attributes
% 4. ICIP paper use age-specific patches
% 5. drawback of the ICIP paper. computational complexity of the Adaboost, location of patches are fixed for every image, and it cannot be trained in a end-to-end manner.
% 6. The new proposed method.
% 7. contributions

\IEEEPARstart{A}{ge-oriented} face synthesis (AOFS) is a generative task aiming to generate older and younger faces by rendering facial images with natural aging and rejuvenating effects. An efficient AOFS method can be integrated into a wide range of forensic and commercial applications, e.g., tracking persons of interest like suspects or missing children over a long time span, predicting the outcomes of a cosmetic surgery, and generating special visual effects on characters of video games, films and dramas~\cite{fu2010age, lanitis2002toward}. The synthesis in recent works \cite{li2019global,wang2018face,yang2018learning,zhang2017age} is usually conducted among age categories (e.g., the 30s, 40s, 50s) rather than specific ages (e.g., 32, 35, 39) since there is no noticeable visual change of a face over a few years. 

\begin{figure}[t]
\begin{center}
\includegraphics[width=0.95\linewidth]{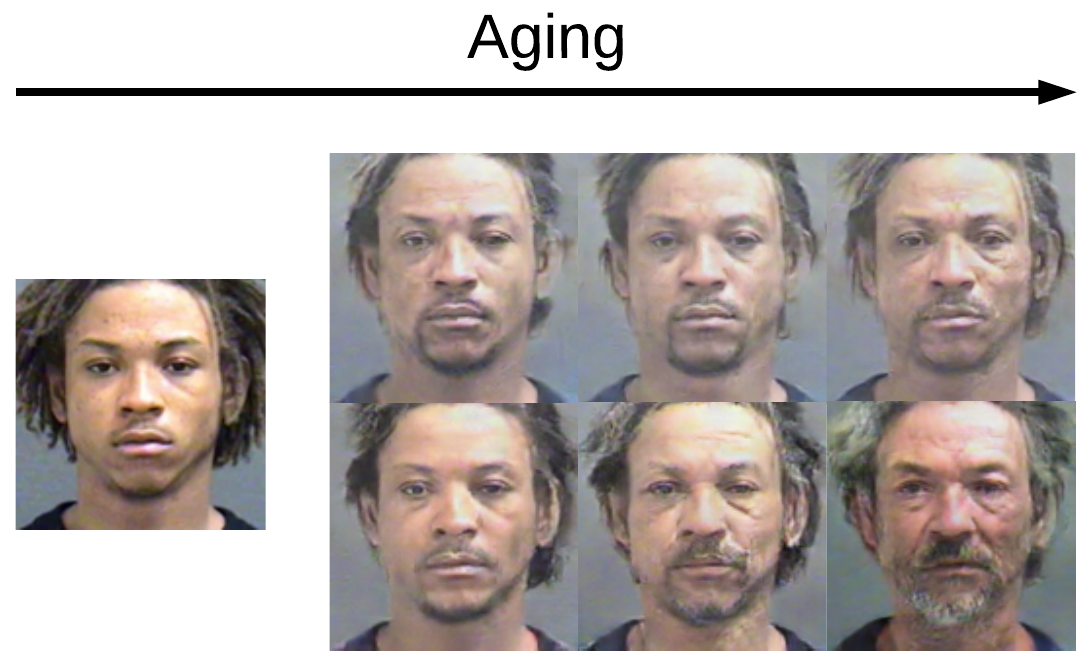}
\end{center}
   \caption{A demonstration of face aging. The top row depict images generated by a vanilla GAN suffering from the mode collapse issue. The bottom row depicts images generated by the proposed AOFS method.}
\label{fig:demo}
\end{figure}

The vanilla Generative Adversarial Network (GAN) \cite{goodfellow2014generative} is commonly used as the backbone of several state-of-the-art AOFS methods \cite{li2019global,antipov2017face,genovese2019towards,pantraki2018face,zhao2019look}. One of the biggest advantages of the vanilla GAN over other generative methods, like the Variational Autoencoder \cite{kingma2013auto}, is that it can generate sharp and realistic images by playing a minimax game between the generator and the discriminator. However, the vanilla GAN suffers from the mode collapse issue caused by the vanishing gradient due to the involvement of the negative log-likelihood loss \cite{arjovsky2017towards}. Specifically, once the discriminator converges, the loss does not penalize the generator any further \cite{che2016mode}. This allows the generator to find a specific mode (i.e., a distribution) that can easily fool the discriminator \cite{bishop2006pattern}. The mode collapse issue may also occur in the AOFS task, where a mode is represented by an age category. Within this context, the vanilla GAN may generate faces with limited variations as exemplified in Fig. \ref{fig:demo}, resulting in poor synthesis accuracy.

To boost the state-of-the-art performance in the AOFS task, this work proposes an AOFS method that includes two novel components. Namely, a Conditional Discriminator Pool (CDP) and an Adversarial Triplet loss. The proposed CDP helps to achieve a high synthesis accuracy by alleviating the mode collapse issue. Specifically, it allows learning multiple modes (i.e., age categories) explicitly and independently to generate realistic faces with a wide range of variations. Our CDP comprises multiple feature-level discriminators that learn the transformations from the source age category to the target age category. For each transformation, only the feature-level discriminator associated with the target age category is used. As a result, each feature-level discriminator only needs to learn one age category throughout the entire training process. The proposed Adversarial Triplet loss helps to preserve the identity information in the synthesized faces. This loss, which improves the Triplet loss [53], uses an additional ranking operation that can further optimize the distances within a triplet of feature embeddings comprising an \textit{anchor}, a \textit{positive} and a \textit{negative}. Specifically, it helps to bring the \textit{positive} much closer to the \textit{anchor}, while guaranteeing that the distance between the \textit{anchor} and the \textit{negative} is larger than that between the \textit{anchor} and the \textit{positive}. The additional ranking operation forces the triplets to a play zero-sum game [5] during training. As a result, our Adversarial Triplet loss yields high-density clusters with dramatically reduced intra-class variances in the feature space.

Our contributions can be summarized as follows. 
\begin{itemize}
\item We study the mode collapse issue in the AOFS task. To the best of our knowledge, our work is the first to tackle the AOFS task from the aspect of mode learning.
\item To address the mode collapse issue in the vanilla GAN and attain a high synthesis accuracy, we propose the CDP, which allows our AOFS method to learn multiple modes explicitly and independently.
\item To preserve the identity information in the synthesized images, we propose the Adversarial Triplet loss. Smaller intra-class variance can be achieved by forcing triplets to play zero-sum games during training.
\item We extensively evaluate the proposed AOFS method on several AOFS benchmark datasets to show that it can precisely transform faces to the target age category while preserving the identity information robustly.
\end{itemize}

The rest of this paper is organized as follows. In Section 2, we review the related works on GANs, especially those tackling the mode collapse issue. In this Section, we also review the Triplet loss and the state-of-the-art AOFS methods. In Section 3, we present details of the proposed AOFS method including the CDP and the Adversarial Triplet loss. In Section 4, we explain the experimental settings and discuss the performance results on several AOFS benchmark datasets. Finally, we conclude our work in Section 5.  

%% ---------------------------------------------------------------------------- %%
%% ---------------------------------------------------------------------------- %%
%% ---------------------------------------------------------------------------- %%

\section{Related Work}

% In this paper, we propose to alleviate the mode collapse issue in the vanilla GAN by using the CSD mechanism and improve the identity permanence of the age-oriented face synthesis method by implementing the proposed DP-Triplet loss.  

To lay the foundation of our work, in this section we first discuss the mode collapse issue in the vanilla GAN and some of the solutions that have been proposed to tackle it. Then, we review the Triplet loss and show the main differences between the proposed Adversarial Triplet loss and other variations. Finally, we discuss some state-of-the-art AOFS methods.

%% ---------------------------------------------------------------------------- %%

\subsection{Mode collapse in GANs}

The vanilla GAN, which is introduced by Goodfellow \textit{et al.} \cite{goodfellow2014generative}, is capable of generating sharp and realistic images by playing a minimax game between its generator and its discriminator. When training the vanilla GAN, the generator and the discriminator try to reach a Nash equilibrium \cite{metz2017unrolled} by minimizing the negative log-likelihood loss and minimizing the JS-divergence \cite{lin1991divergence}. However, the involvement of the negative log-likelihood loss may cause the discriminator to converge faster than the generator \cite{heusel2017gans}. Once the discriminator finds its global minima, the loss function stops penalizing the generator \cite{che2016mode}. This is also known as the vanishing gradient problem \cite{arjovsky2017towards,fedus2017many,jolicoeur2019relativistic} and is the main cause of the mode collapse issue. Since the parameters in the discriminator are not further updated, the generator may then find a specific mode that can easily fool the discriminator. When such an issue occurs, the vanilla GAN can only generate limited varieties of samples. Solving this mode collapse issue has become one of the most trending research topics on GANs.

Since the mode collapse issue is caused by the vanishing gradient problem due to the involvement of the negative log-likelihood loss, one strategy to alleviate it is to use an alternative loss function that minimizes a different divergence. Nowozin \textit{et al.} \cite{nowozin2016f} first show that the optimization of GANs is a general process that can be done by minimizing any $f$-divergence \cite{csiszar2004information,liese2006divergences}, which is a family of divergences aiming to minimize the distance between two distributions. Some commonly used members of the $f$-divergence family are the JS-divergence, the Kullback-Leibler divergence (KL-divergence) \cite{kullback1951information}, the squared Hellinger divergence, and the Pearson $\chi^2$ divergence \cite{pearson1900x}. The authors show that GANs trained with other divergences, like the KL-divergence or the squared Hellinger divergence, can generate images with more variations compared to those generated by the vanilla GAN. Although the work in \cite{nowozin2016f} does not tackle the mode collapse issue directly, it shows the possibility of using other loss functions to optimize GANs.

Arjovsky \textit{et al.} \cite{arjovsky2017wasserstein} propose the Wasserstein GAN (WGAN) and use the Wasserstein or Earth-Mover (EM) distance to calculate the distance between distributions of the real and synthesized data. Intuitively, the EM distance computes the cost of transforming one distribution to another, which is more sensitive to the difference between two distributions \cite{arjovsky2017wasserstein}. Therefore, even if the discriminator is well-trained, it can still keep rejecting the data synthesized by the generator. The Least Square GAN (LSGAN) \cite{mao2017least}, on the other hand, replaces the negative log-likelihood loss by the L1 loss. Minimizing the L1 loss is equivlent to minimizing the Pearson $\chi^2$ divergence, which can produce overdispersed approximations and thus makes the LSGAN less mode-seeking \cite{dieng2017variational, mao2018effectiveness}. 

Although the methods discussed before may alleviate the mode collapse issue, their discriminators still have to learn from all the modes. Therefore, recently proposed methods now focus on modifying the GAN structure. For example, Nguyen \textit{et al.} \cite{nguyen2017dual} propose the Dual Discriminator Generative Adversarial Nets (D2GAN) where each discriminator favors data from a different distribution. By using this strategy, their method can compute the KL and reverse KL divergence simultaneously, which in turn increases the variety of samples. Based on this idea, Zhang \textit{et al.} \cite{zhang2018convergence} propose a D2GAN variation with two customized discriminators. Specifically, one discriminator consists of residual blocks to form a deep network aiming to increase the variety of generated samples. The other discriminator uses the scaled exponential linear unit (SELU) function \cite{klambauer2017self} as the non-linear activation function. Adopting the SELU function guarantees that this discriminator produces a non-zero value even if the distributions of the synthesized and real data are similar. The authors further propose the D2PGGAN \cite{zhang2019d2pggan} to stabilize the training by leveraging the idea of progressively increasing the complexity of the generator \cite{karras2018progressive}. Durugkar \textit{et al.} \cite{durugkar2016generative} propose a GAN with multiple discriminators. Their method may alleviate the mode collapse issue to some extent since the generator has to fool a set of discriminators, which in turn makes the generated samples diverse. It is important to note that by introducing additional discriminators in parallel, the aforementioned methods are also more computationally complex than their plain counterparts (e.g., the vanilla GAN). On the contrary, by selecting a particular discriminator from a discriminator pool, our CDP only uses one discriminator for each transformation, which does not increase the computational complexity. 

\begin{figure*}
\begin{center}
\includegraphics[width=0.98\textwidth]{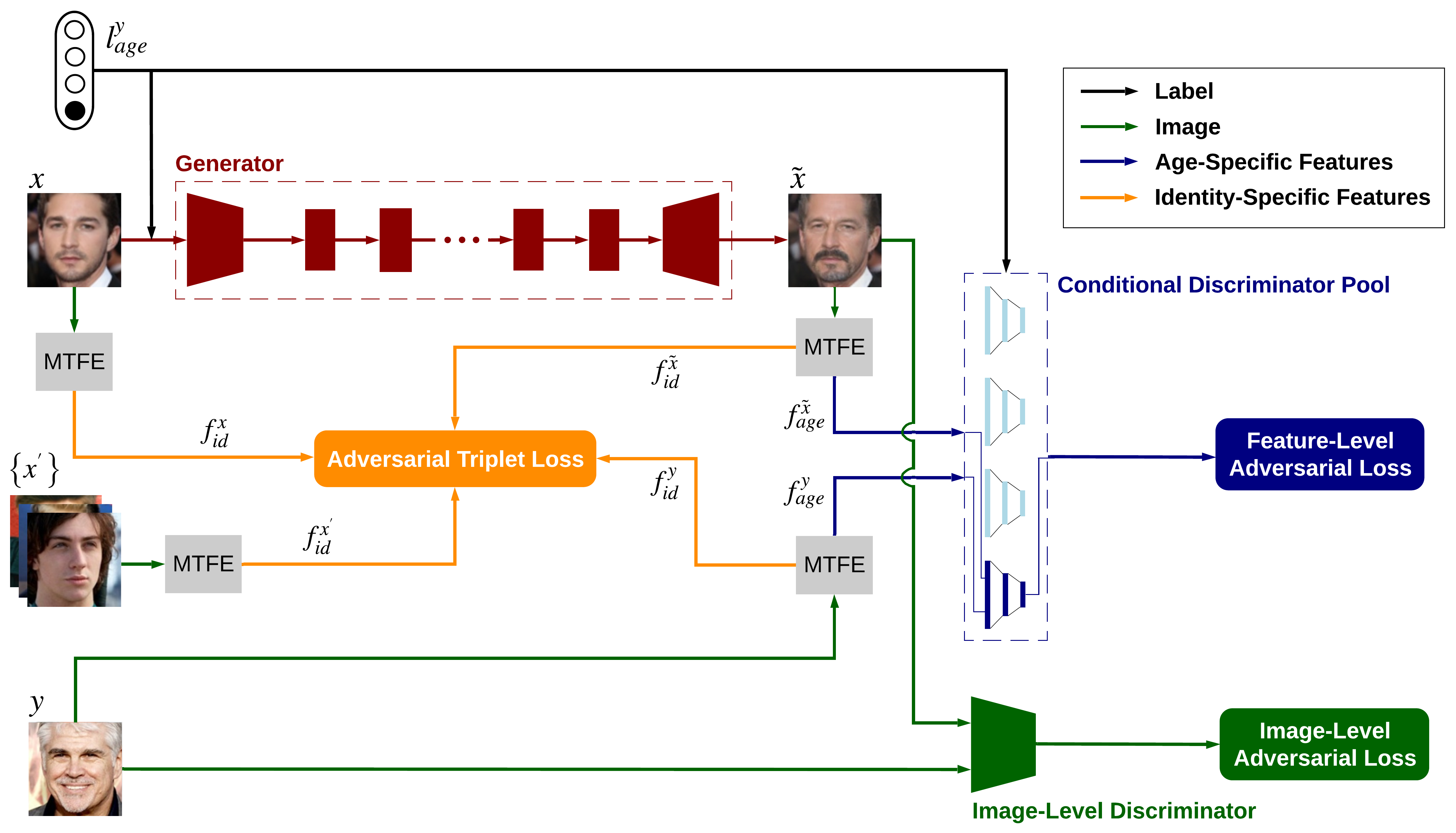}
\end{center}
   \caption{Architecture of the proposed AOFS method. It consists of a generator with residual blocks (red rectangles), an image-level discriminator, and a CDP that contains several feature-level discriminators. The number of feature-level discriminators equals the number of age categories that the method should learn. Two adversarial losses are used to synthesize realistic aged and rejuvenated faces. To further optimize the identity features in the synthesized image, $\tilde{x}$, we leverage additional input images, $\left \{x'\right \}$, that are within the same age category as the source image, $x$. Image $y$ carries the target age information for $\tilde{x}$.}
\label{fig:architecture}
\end{figure*}

%% ---------------------------------------------------------------------------- %%

\subsection{Triplet Loss}

% To preserve the identity information on synthesized faces, state-of-the-art AOFS methods \cite{genovese2019towards,liu2019attribute,wang2018face,yang2018learning,zhao2019look} use the L1 constraint to minimize the distance between two identity-specific features from the source image and the synthesized image. It is no doubt that the L1 constraint is intuitive and easy to compute. However, the aforementioned methods all assign a trivial weighting parameter to this identity preserving loss compared to other losses used in their method, which implicitly limits the identity permanence ability of these AOFS methods. Moreover, increasing the weighting parameter will distort the synthesized images, resulting in a poor synthesis accuracy. To avoid this problem and efficiently preserve the identity information in synthesized images, we use the proposed DP-Triplet loss in our AOFS method.

The Triplet loss is proposed in \cite{schroff2015facenet} aiming to learn feature embeddings for images by optimizing the geometric relationship, in the feature space, within a triplet consisting of an \textit{anchor}, a \textit{positive} and a \textit{negative}. Within this context, the \textit{anchor} and \textit{positive} represent feature embeddings of the same class and the \textit{negative} represents a feature embedding of a different class. The goal is to minimize the distance between the \textit{anchor} and the \textit{positive} and simultaneously push the \textit{negative} away from the \textit{anchor}. Since then, a number of variations to this loss have been proposed. For instance, Chen \textit{et al.} \cite{chen2017beyond} uses an additional \textit{negative} embedding alongside the original triplet to form a quadruplet. Huang \textit{et al.} \cite{huang2016learning} implement three ranking operations in total by using an \textit{anchor}, a \textit{negative} and three \textit{positives}. Ye \textit{et al.} \cite{ye2018visible}, on the other hand, adopt additional images from other modalities. It is worth noting that all these variants leverage additional samples either within the same or from another modality. Therefore, these losses can no longer help to optimize the geometric relationship within a triplet. This is explained in detail in Section III.D.

We find that the original Triplet loss produces clusters with large intra-class variances that can be further optimized. To produce high-density clusters, we add another ranking operation and propose the Adversarial Triplet loss to pull the \textit{positive} closer to the \textit{anchor}. It is worth noting that compared to the aforementioned Triplet loss variants, our Adversarial Triplet loss still focuses on optimizing distances within triplets without leveraging additional samples.

%% ---------------------------------------------------------------------------- %%

\subsection{Age-Oriented Face Synthesis}

The first AOFS methods can be traced back to~\cite{mark1983perception,mark1981perception,todd1980perception}, in which craniofacial growth in young faces is studied. In the early stage, geometry-based methods were a popular choice among researchers, and one of the most representative works is the Active Shape Model (ASM) \cite{cootes1995active}. The authors model the shape of faces by adjusting the positions of a number of points. Each point marks one part of the face, such as the position of the eyes and the boundary of the face. Synthetic facial images of different shapes and ages can then be obtained by adjusting the position of these points. Another approach to rendering aging or rejuvenating effects is to directly synthesize or remove wrinkles on a given facial image \cite{bando2002simple,liu2004image,mukaida2004extraction,wu1999simulating,wu1995dynamic}. Later, Ramanathan and Chellappa \cite{ramanathan2006modeling} propose an aging-focused method called the craniofacial growth model for synthesizing elderly faces by leveraging facial landmark movements. Another worth-noting early AOFS method is \cite{tang2018personalized}, where the authors use dictionary learning to learn a personalized aging process, and associate an aging dictionary to each subject to represent their aging characteristics.

With the increasing popularity of deep learning, several attempts have been made to tackle the AOFS problem using various network architectures. Both Wang \textit{et al.} \cite{wang2018face} and Zhang \textit{et al.} \cite{zhang2017age} use conditional adversarial learning \cite{mirza2014conditional} to synthesize aged faces. Wang \textit{et al.} further employ an age category classifier to boost the synthesis accuracy and an L2 constraint on the identity-specific features to preserve the identity information. Yang \textit{et al.}~\cite{yang2018learning} propose a GAN framework by implementing a customized discriminator with a pyramid architecture, which leads to more realistic results than a conventional discriminator as images can be discriminated based on features at multiple scales. They further adopt a pre-trained identity classifier to preserve the identity in the synthesized images. AOFS methods based on the Wavelet transform are proposed recently in \cite{li2019global,liu2019attribute}, where this transform is used to enhance the texture information in the frequency domain so that richer aging and rejuvenating effects can be synthesized. He \textit{et al.} \cite{he2019s2gan} implement a GAN model with a customized generator, where a number of decoders are implemented, each one learning an age category. All the decoders are associated with a weight factor to control their relative importance in each transformation. Since all the decoders are trained in parallel, the computational complexity of the method is proportional to the number of age categories to be learned.

Our work is different from the aforementioned deep-learning methods as it tackles the AOFS problem from a different angle, i.e., mode learning. Our method can achieve a high synthesis accuracy by learning multiple modes explicitly and independently. Additionally, compared to the L1 loss, the L2 loss, and the simple classifiers used in the those methods, our AOFS method uses the proposed Adversarial Triplet loss to keep the identity information unaltered in the synthesized facial images.

%% ---------------------------------------------------------------------------- %%
%% ---------------------------------------------------------------------------- %%
%% ---------------------------------------------------------------------------- %%

\section{Proposed AOFS Method}
In this section, we explain in detail our proposed method by first formulating the problem and explaining the pre-trained Multi-Task Feature Extractor (MTFE) used to extract age-specific and identity-specific features. We then present the proposed CDP and the Adversarial Triplet loss. Finally, we explain the overall loss used to train our method. 

%% ---------------------------------------------------------------------------- %%

\subsection{Problem Formulation}
Since the transformation is conducted among age categories rather than specific ages, following the prior work in \cite{li2019global,yang2018learning,liu2019attribute}, we divide the data into four categories according to the following age ranges: $30^-$, $31 - 40$, $41 - 50$, and $51^+$. Each category is denoted by $C_i$, where $i\in[1,4]$.

To render aging and rejuvenating effects, the proposed AOFS method takes two faces, $x\in{C_X}$ and $y\in{C_Y}$, and the age label of $\mathbf{\mathit{y}}$, $l_{age}^y$, as the inputs, where ${X}\neq{Y}$. Specifically, $x$ is the face that is to be aged or rejuvenated and $y$ carries the desired age information. Our method aims to generate an aged or rejuvenated $x$, denoted by $\tilde{x}$, which is expected to belong to the same age category as $y$. Moreover, to ensure that the identity information is effectively preserved in $\tilde{x}$, our method also uses other images in the same batch, $\left \{x'\right \}$, to compute the Adversarial Triplet loss. It is worth noting that both $x'$ and $y$ do not share the same identity information of $x$.

% For each transformation, the proposed AOFS method takes two faces, $x\in{C_X}$ and $y\in{C_Y}$, and the age label of $\mathbf{\mathit{y}}$, $l_{age}^y$, as the inputs, where ${X}\neq{Y}$ and $\mathbf{\mathit{y'}}$ is only used to compute the Adversarial Triplet loss. Specifically, $\mathbf{\mathit{x}}$ is the face that is to be aged or rejuvenated, $\mathit{y}$ carries the desired age information, and $\mathit{y'}$ shares the same identity with $\mathit{y}$. Our model aims to generate an aged or rejuvenated $\mathbf{\mathit{x}}$, denoted by  $\mathbf{\mathit{\tilde{x}}}$, which is expected to belong to the same age category as $\mathbf{\mathit{y}}$. Moreover, the identity information after each transformation is expected to remain unaltered, i.e. $l_{id}^x = l_{id}^{\tilde{x}}$. Note that for each ${\mathbf{\mathit{x}}}$, by using a different ${\mathbf{\mathit{y}}}$ and corresponding $l_{age}^y$, our model can generate aged and rejuvenated faces simultaneously.

In summary, the proposed method achieves three goals simultaneously: 1) To generate realistic aged and rejuvenated faces; 2) to force the synthesized faces to be within the target age category; and 3) to preserve the identity information in the synthesized image. The architecture of our proposed AOFS method is illustrated in Fig. \ref{fig:architecture}.

%% ---------------------------------------------------------------------------- %%

\subsection{Multi-Task Feature Extractor}
\begin{figure}[t]
\begin{center}
\includegraphics[width=1\linewidth]{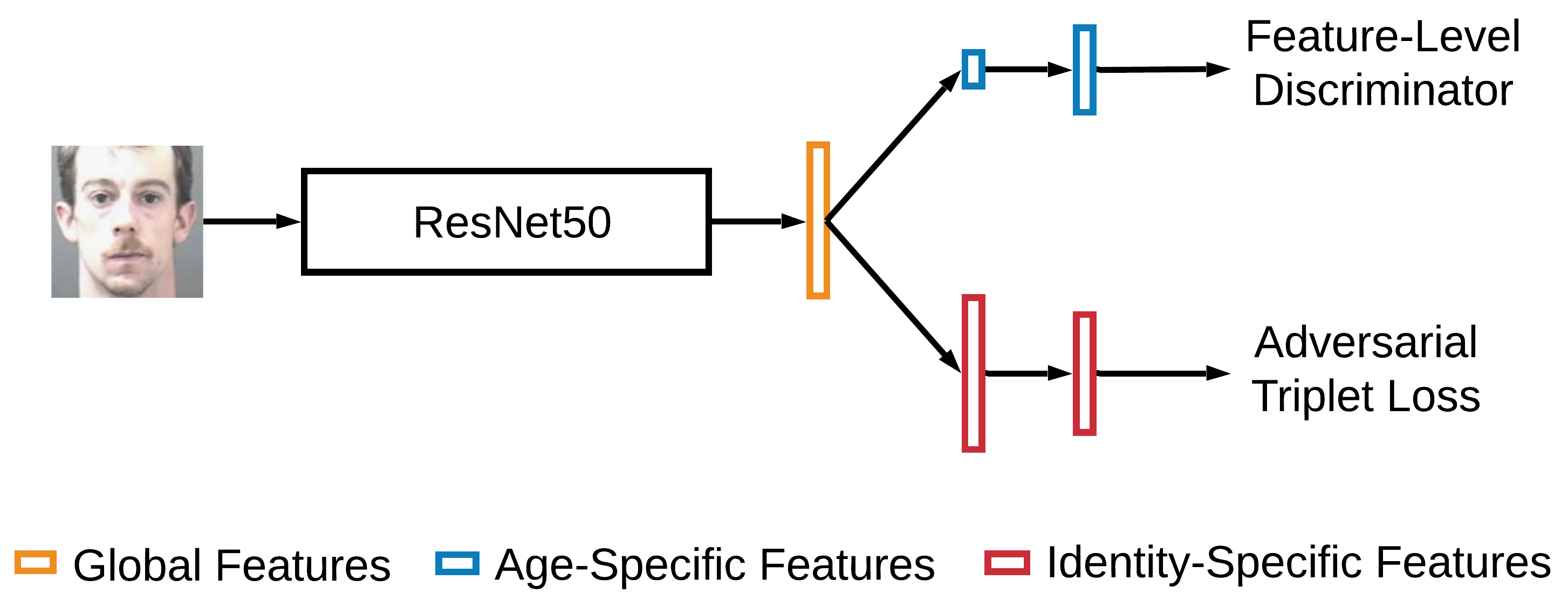}
\end{center}
   \caption{Architecture of our MTFE. After the decomposition, we resize each set of task-specific features to be used by the corresponding feature-level discriminator of the CDP or the Adversarial Triplet loss.}
\label{fig:fx}
\end{figure}

The CDP and the Adversarial Triplet loss of the proposed AOFS method use age-specific and identity-specific features from input images and synthesized images. To extract and disentangle these features, we use the decomposition method proposed in \cite{wang2018orthogonal}. Specifically, we use a ResNet-50 \cite{he2016deep} as the backbone. The architecture of this feature extractor is depicted in Fig. \ref{fig:fx}. This model decomposes all the features extracted from a facial image into two components based on a spherical coordinate system, which is formulated as:
\begin{equation}\label{decomposition}
    {f}_{sphere} := \left \{r; \textbf{\textit{theta}} \right \},
\end{equation}
where the ${f}_{sphere}$ is the set of features after the decomposition in which the angular component \textbf{\textit{theta}} = $\left \{\theta_1, \theta_2, ..., \theta_k \right \} $ indicate the identity-specific features for $k$ identities, and the radial component $r$ encodes the age-specific features. 

We replace the regression loss used to learn age-specific features in \cite{wang2018orthogonal} with an age regression model~\cite{rothe2018deep,wang2018fusion} to supervise the age-specific learning process, which has been shown to achieve better performance for the age estimation task. We observe that feature extractors trained in this multi-tasking manner can achieve higher accuracy on both the age category classification and identity classification tasks than single-task networks. Additionally, we use our proposed Adversarial Triplet loss to learn identity-specific features.

%% ---------------------------------------------------------------------------- %%

\subsection{Conditional Discriminator Pool}
In the vanilla GAN with a single image-level discriminator, the loss function for face synthesis is usually formulated as:
\begin{equation}\label{image_adversarial}
\begin{aligned}
    % \mathcal{L}_{adv} = &\mathbb{E}_{x\sim P_{x}}[logD(x)] \\
    %                         &+ \mathbb{E}_{z\sim P_{z}}[log(1-D(G(z))],
    \mathcal{L}_{adv} = &\mathbb{E}_{y}[logD(y)] \\
                            &+ \mathbb{E}_{x}[log(1-D(G(x))],
\end{aligned}
\end{equation}
where $G$ is the generator trying to minimize the loss, and $D$ is the discriminator trying to maximize the loss. As mentioned before, GANs based on this loss function suffer from the mode collapse issue. To force the network to learn each mode independently and thus alleviate this issue, one can add more discriminators directly. However, such an strategy may lead to a high computational complexity and redundancy during training, as not all the discriminators are expected to back-propagate the loss during each transformation. Therefore, we propose a mechanism to select the corresponding discriminator for each transformation based on the input label that represents the target age information. Let us recall that our proposed AOFS method treats each age category as a mode, which results in four modes in total. We use the input label, $l_{age}^y$, to select the corresponding discriminator that learns the target age category. Our proposed method implements this mechanism on discriminators at the feature level, which are used to synthesize aging and rejuvenating effects. Therefore, we assemble four feature-level discriminators with an identical architecture to form our CDP. Each feature-level discriminator targets one mode. Our method additionally uses an image-level discriminator to remove artificial effects from the synthesized faces. As illustrated in Fig. \ref{fig:architecture}, in each transformation, our method leverages the selected feature-level discriminator alongside the image-level discriminator.

It is important to note that an alternative way to select the feature-level discriminator is by employing an additional classifier. However, within the context of AOFS, the accuracy of classifying age categories may be very low, from 25\% to 60\% depending on the specific age category in different AOFS benchmark datasets \cite{wang2018face,liu2019attribute}. Employing such a low-accuracy classifier may result in a selecting a discriminator that learns an incorrect mode. Instead, we directly use $l_{age}^y$ to select discriminators, which guarantees that, in each transformation, the discriminator associated with the target mode is used. We then formulate the feature-level adversarial loss as follows:
\begin{equation}\label{feature_adversarial}
\begin{aligned}
    \mathcal{L}_{adv_{feature}} = \mathbb{E}_{f_{age}^y}[log(FD_{C_i}(f_{age}^y)|l_{age}^y)] \\
                                 + \mathbb{E}_{f_{age}^{\mathit{\tilde{x}}}}[log(1-(FD_{C_i}(f_{age}^{G(x|l_{age}^y)})|l_{age}^y))],
\end{aligned}
\end{equation}
where $FD_{C_i}$ is the selected feature-level discriminator trying to maximize the loss; $f_{age}^y$ denotes the age-specific features extracted from the target image, $y$; and $f_{age}^{G(x|l_{age}^y)}$ denotes the age-specific features extracted from the synthesized image, $\tilde{x}$, where $G(x|l_{age}^y)$ is the generator that produces $\tilde{x}$ conditioned on $l_{age}^y$. Finally, $l_{age}^y$ is a one-hot encoded vector indicating the label for the target age category, $C_i$.

\begin{figure*}
\begin{center}
\includegraphics[width=0.98\textwidth]{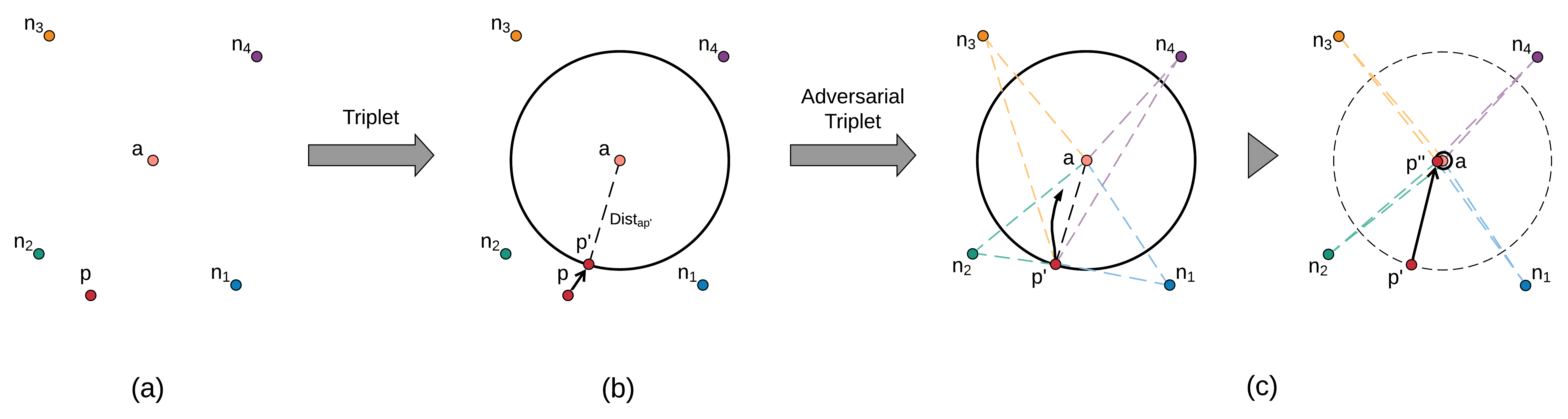}
\end{center}
   \caption{An example showing how the Adversarial Triplet loss works. $a$ (\textit{anchor}) and $p$ (\textit{positive}) are feature embeddings representing the same class. The \textit{negatives} $n_1$, $n_2$, $n_3$, and $n_4$ indicate feature embeddings from other classes, each one from a distinct class. 
   (a) Original positions of these feature embeddings. (b) By using the Triplet loss, $p$ can move towards $p'$ when minimizing Eq. (\ref{triplet}). (c) Our Adversarial Triplet loss guarantees that for each $n_i$ where $i \in [1,2,3,4]$, $Dist_{an_i} \approx Dist_{n_{i}p}$ by adding an additional operation as formulated in Eq. (\ref{adversarial-triplet}). In this case, $p'$ may continue moving towards $a$ and end up at a location which is extremely close to it, i.e., $p''$.}
\label{fig:adversarial-triplet}
\end{figure*}

%% ---------------------------------------------------------------------------- %%

\subsection{Adversarial Triplet Loss}
The Triplet loss~\cite{schroff2015facenet} with three feature embeddings is formulated as:
\begin{equation}\label{triplet}
\begin{aligned}
    \mathcal{L}_{Triplet} (a,p,n) =& \sum_{a,p,n} [m + Dist_{a,p} - Dist_{a,n}]_+,
\end{aligned}
\end{equation}
where $Dist_{j,k}$ indicates the Euclidean distance between embeddings $j$ and $k$ in the feature space and $a, p, n$ are the indices of the \textit{anchor}, the \textit{positive} and the \textit{negative}, respectively. This loss forces $Dist_{a,n}$ to be larger than $Dist_{a,p}$ by at least a margin $m$. However, once this criterion is satisfied, $Dist_{a,p}$ cannot be further minimized, which may lead to large intra-class variances. To overcome this problem, we add another ranking operation to Eq. (\ref{triplet}), which forces $Dist_{a,n}$ to be larger than the distance between $n$ and $p$, $Dist_{n,p}$. This additional operation helps to further bring $p$ closer to $a$ by forcing different triplets with the same $a$ and $p$ but different $n$ to play a zero-sum game: 
\begin{equation}\label{adversarial-triplet}
\begin{aligned}
    \mathcal{L}_{AT} (a,p,n) = & \sum_{a,p,n} [m + Dist_{a,p} - Dist_{a,n}]_+ \\
    &+ [Dist_{n,p} - Dist_{a,n}].
\end{aligned}
\end{equation}

Let us assume there are several triplets with the same $a$ and $p$, but different $n$, where each distinct $n$ is denoted by $n_i$. Under this assumption, the Triplet loss in Eq. (\ref{triplet}) can be minimized as long as $Dist_{a,n_{i}} > Dist_{a,p} + m$, which may result in clusters with large intra-class variances. To reduce such variances, $Dist_{a,n_{i}}$ should be larger than $Dist_{n_{i},p}$. Let us take the triplets $a-p-n_1$ and $a-p-n_3$ in Fig. \ref{fig:adversarial-triplet} as an example, where $n_1$, $n_2$, $n_3$, and $n_4$ are all from different classes. In this example, both $n_1$ and $n_3$ should maintain their relative position with respect to the $a-p$ cluster in order to also be far from other neighboring clusters. In other words, $n_1$ and $n_3$ should not move towards either $n_2$ or $n_4$. In this case, $\mathcal{L}_{AT} (a,p,n_1)$ tries to pull $p$ towards $n_1$ and minimize $Dist_{n_{1},p}$, while $\mathcal{L}_{AT} (a,p,n_3)$ tries to pull $p$ towards $n_3$ and minimize $Dist_{n_{3},p}$. Therefore, $\mathcal{L}_{AT} (a,p,n_1)$ and $\mathcal{L}_{AT} (a,p,n_3)$ play a zero-sum game as minimizing one loss increases the other. This is also true for $\mathcal{L}_{AT} (a,p,n_2)$ and $\mathcal{L}_{AT} (a,p,n_4)$. In order to minimize all losses in this example, i.e., to have a total loss equal to zero, $p$ should be in the same position as $a$ so that $Dist_{a,n_{i}} = Dist_{n_{i},p}$. In practice, however, our Adversarial Triplet loss pulls $p$ to a position very close to $a$ so that $Dist_{a,n_{i}} \approx Dist_{n_{i},p}$.

\begin{figure}
\begin{center}
\includegraphics[width=1\linewidth]{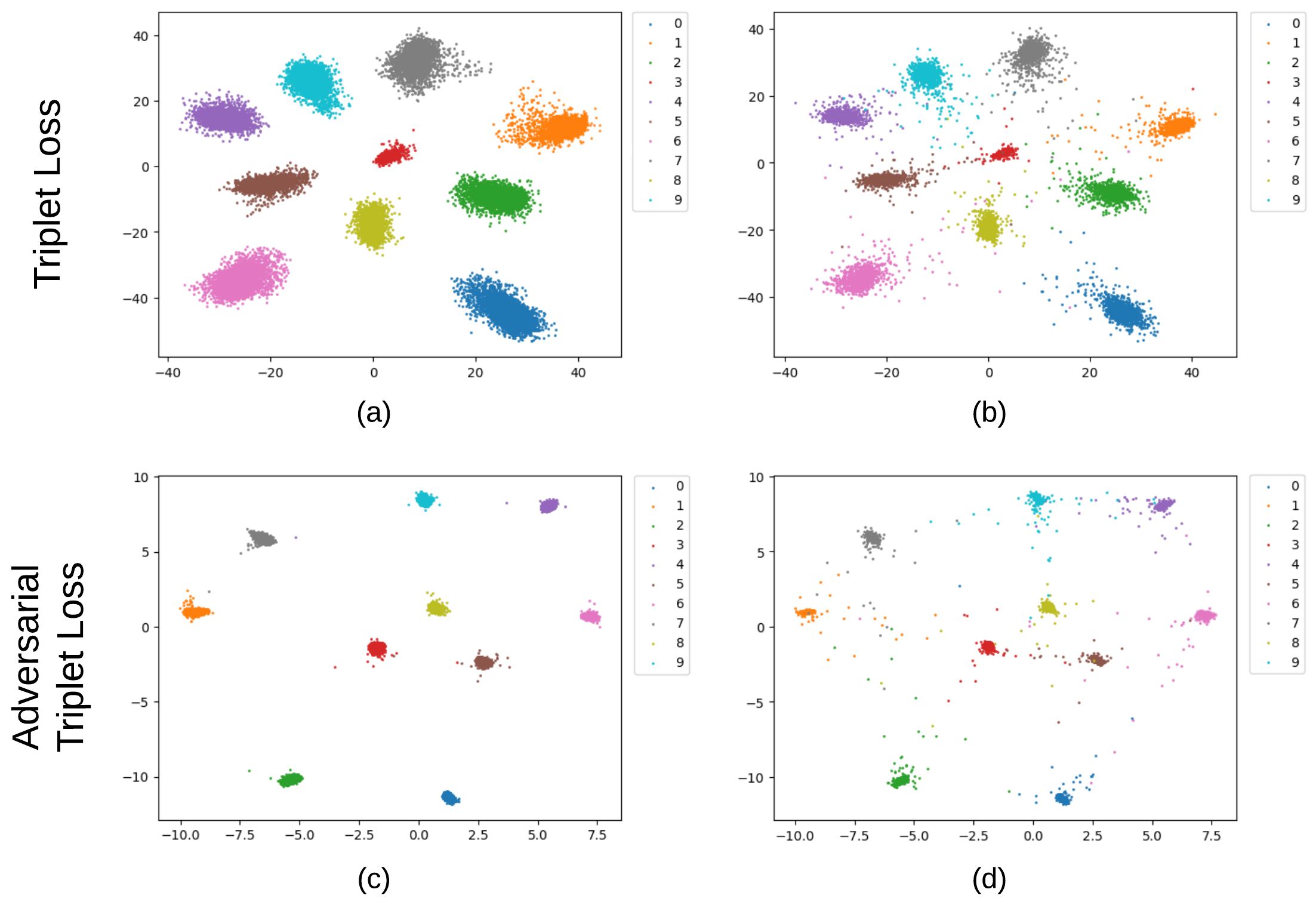}
\end{center}
   \caption{Feature distribution of the MNIST dataset for classification on (a),(c) the training set and (b),(d) the test set when the Triplet loss and the Adversarial Triplet loss are used.}
\label{fig:triplet_mnist}
\end{figure}

Fig. \ref{fig:triplet_mnist} demonstrates the performance of the Adversarial Triplet loss on a real dataset. In this example, the feature distribution of the MNIST dataset for classification is presented. To this end, we employ an Alexnet \cite{krizhevsky2012imagenet} as the deep network, but replace all the fully-connected layers, except the output layer, by a single linear layer with two neurons for visualization purposes. From the figure, we can observe that the features learned by the Adversarial Triplet loss dramatically reduce the intra-class variances compared to the features learned by the Triplet loss. The classification accuracy attained by each loss is tabulated in Table \ref{tab:triplet_mnist}.

% In a conventional scenario like training the feature extractor and on the MNIST dataset, the \textit{anchor} in one triplet could be \textit{positive} or \textit{negative} in others. With enough training iterations, each embedding can act as \textit{negative} in some triplets and is pushed two times away from its surrounding clusters based on two ranking operations in the above equation. Therefore, during the entire training process, all embeddings belong to one cluster are moving towards a point that is far away from all neighboring clusters, which leads to high-density clusters.

One of the most critical issues in the Triplet loss is that as the number of triplets grows, many triplets can easily satisfy the constraint in Eq. (\ref{triplet}), which in turn may lead to poor convergence \cite{schroff2015facenet}. To overcome this issue in the Adversarial Triplet loss, we adopt a hard negative mining strategy \cite{hermans2017defense}. Specifically, we use an online hard sample mining method in which each batch consists of samples from $T$ classes, and each class has $S$ samples within one batch, for a batch size of $B=TS$. In this method, each sample in a batch acts as the \textit{anchor} for one triplet, thus, there are a total of $B$ triplets within one batch. For each \textit{anchor}, a hardest \textit{positive} sample with the largest distance and a hardest \textit{negative} sample with the smallest distance are selected to form a triplet. This method does not require pre-defining the triplets and can generate hard triplets in an online manner. After incorporating this hard sample mining strategy, our Adversarial Triplet loss in Eq. (\ref{adversarial-triplet}) is as follows:
\begin{equation}\label{adversarial-triplet_modified}
\begin{aligned}
    \mathcal{L}_{AT} (a,p,n) = &\sum_{\substack{t=1}}^T\sum_{\substack{s=1}}^S[m + \max_{\substack{p}}Dist_{a,p} - \min_{\substack{n}}Dist_{a,n}]_+ \\
    &+ [Dist_{n,p} - \min_{\substack{n}}Dist_{a,n}],
\end{aligned}
\end{equation}
where $t$ is the class index and $s$ is the image index for each class in one batch. 

\begin{table}
\ra{1.05}
\begin{center}
\caption{Classification accuracy (\%) n on the MNIST dataset.}
\label{tab:triplet_mnist}
\begin{tabular}{l@{\hskip 0.2in}c@{\hskip 0.2in}c}\toprule
\hfil{Loss} & \hfil{Triplet} & \hfil{Adversarial Triplet} \\ \midrule
\hfil{Accuracy} & \hfil{99.43} & \hfil{\textbf{99.67}} \\
\bottomrule
\end{tabular}
\end{center}
\end{table}

Since we are trying to optimize the identity-specific features on the synthesized faces when training our AOFS method, we use the identity-specific features, $f_{id}^{x}$, from the source image as the \textit{anchor} and the identity-specific features, $f_{id}^{\tilde{x}}$, from the synthesized image as the \textit{positive}. In addition, we use all other images in the same batch that do not share the same identity with the source image as the \textit{negatives}. The Adversarial Triplet loss of our AOFS method with the hard sample mining strategy is then formulated as:  
\begin{equation}\label{adversarial-triplet_AOFS}
\begin{aligned}
    \mathcal{L}_{AT} &(f_{id}^{x},f_{id}^{\tilde{x}},\left \{ f_{id}^{x'},f_{id}^{y} \right \}) = \sum_{\substack{t=1}}^T\sum_{\substack{s=1}}^S \\
    &[m + Dist_{f_{id}^{x},f_{id}^{\tilde{x}}} - \min_{\substack{\left \{f_{id}^{x'},f_{id}^{y}\right \}}}Dist_{f_{id}^{x},\left \{f_{id}^{x'},f_{id}^{y}\right \}}]_+ \\
    &+ [Dist_{\left \{f_{id}^{x'},f_{id}^{y}\right \},f_{id}^{\tilde{x}}} - \min_{\substack{\left \{f_{id}^{x'},f_{id}^{y}\right \}}}Dist_{f_{id}^{x},\left \{f_{id}^{x'},f_{id}^{y}\right \}}],
\end{aligned}
\end{equation}
where $\left \{f_{id}^{x'}\right \}$ are the identity-specific features of images within the same age category as the source image but carrying different identity information, and $f_{id}^{y}$ are the identity-specific features of images within the target age category. It is worth noting that the above equation do not have the $max$ operation as in Eq. (\ref{adversarial-triplet_modified}) since the \textit{positive} in this case, $f_{id}^{\tilde{x}}$, is synthesized thus cannot be selected.

\subsection{Overall Loss}
The image-level adversarial loss in our AOFS method is formulated as:
\begin{equation}\label{feature_adversarial}
\begin{aligned}
    \mathcal{L}_{adv_{image}} = &\mathbb{E}_{y}[logD(y)] \\
                            &+ \mathbb{E}_{x}[log(1-D(G(x|l_{age}^y))].
\end{aligned}
\end{equation}

The overall loss function, $\mathcal{L}_{overall}$, to train our method is a weighted summation of several losses, with $\mathcal{L}_{adv_{image}}$ removing ghost artifacts, $\mathcal{L}_{adv_{feature}}$ synthesizing ageing and rejuvenating effects and attaining a high synthesis accuracy, and $\mathcal{L}_{AT}$ preserving the identity information:

\begin{equation}\label{overall}
\begin{aligned}
    \mathcal{L}_{overall} =& \mathcal{L}_{adv_{image}} + \lambda_{adv_{feature}}\mathcal{L}_{adv_{feature}} \\
    &+ \lambda_{AT}\mathcal{L}_{AT},
\end{aligned}
\end{equation}
where $\lambda_{adv_{feature}}$ and $\lambda_{AT}$ control the relative importance among learning objectives.

\begin{figure*}
\begin{center}
\includegraphics[width=18.4cm]{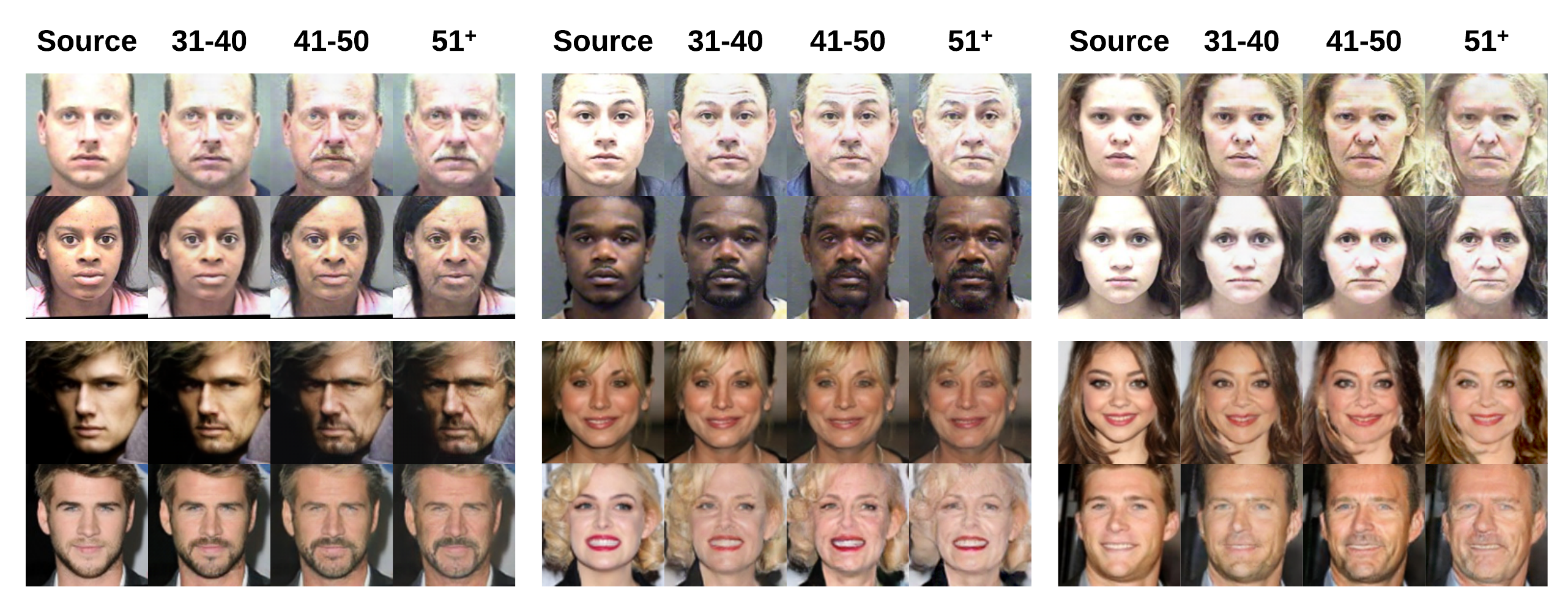}
\end{center}
   \caption{Aging results. The top five rows show the synthesized results on the MORPH II dataset, and the bottom five rows show the synthesized results on the CACD.}
\label{fig:progress_aging}
\end{figure*}

\begin{figure*}
\begin{center}
\includegraphics[width=18.4cm]{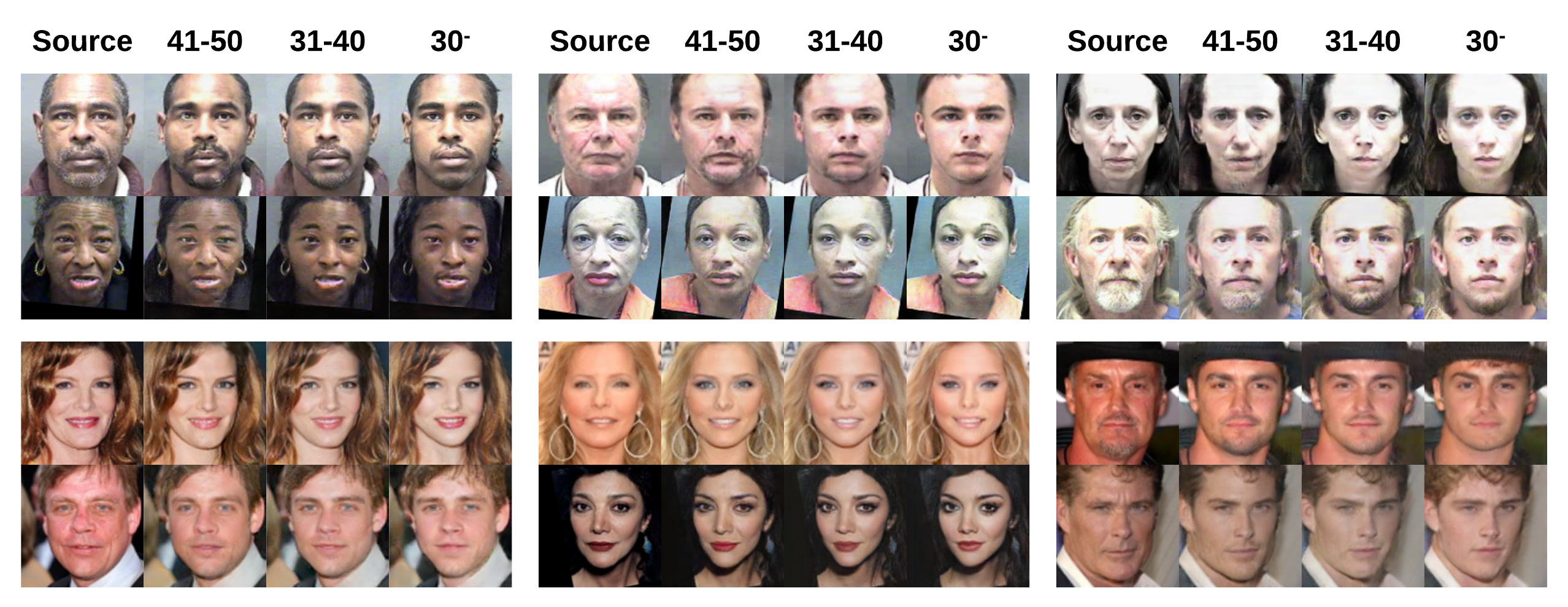}
\end{center}
   \caption{Rejuvenating results. The top five rows show the synthesized results on the MORPH II dataset, and the bottom five rows show the synthesized results on the CACD.}
\label{fig:progress_rejuvenating}
\end{figure*}

%% ---------------------------------------------------------------------------- %%
%% ---------------------------------------------------------------------------- %%
%% ---------------------------------------------------------------------------- %%

\section{Experiments}
In this section, we first briefly describe the two AOFS benchmark datasets used in our experiments followed by the implementation details of our method. Then, we compare our method with state-of-the-art methods and conduct ablation studies, both qualitatively and quantitatively, to show that our method can achieve a high synthesis accuracy while preserving the identity information on the synthesized facial images.

%% ---------------------------------------------------------------------------- %%

\subsection{AOFS benchmark datasets}
We use the MORPH II dataset \cite{ricanek2006morph} and the Cross-Age Celebrity Dataset (CACD) \cite{chen2014cross} to train the MTFE and evaluate our method. The MORPH II dataset contains about 55,000 facial images of individuals with ages ranging from 16 to 77. The CACD contains more than 160,000 facial images of individuals with ages ranging from 16 to 62. Most of the images in the MORPH II dataset are mugshots, while images in the CACD contain Pose, Illumination, and Expression (PIE) variations. Each image in both datasets is associated with an age label and an identity label.

% We use the open-source computer vision library dlib~\cite{dlib09} for all image preprocessing. 

All images are cropped to 128$\times$128 pixels and aligned based on the location of the eyes. Since not all images can be aligned by using this technique, in the end, 55,062 images from the MORPH II dataset and 159,226 images from the CACD are used in our experiments. For each dataset, we use 80\% of the images for training and the remaining 20\% for testing. The number of training images for each age category in the MORPH dataset is 19,949, 12,496, 8,982, and 2,622, for the categories $\left \{30^-, 31-40, 41-50, 51^+ \right \}$, respectively. For the CACD, the number of training images of each age category is 39,416, 33,742, 30959, and 23,262, respectively. There is no identity overlap between the training and test sets.

We conduct a five-fold cross validation for all our experiments. For the MORPH II dataset, each fold has about 2,550 subjects with 3,989, 2,499, 1,796, and 524 images within each age category, respectively. For the CACD, each fold contains about 400 subjects with 7,883, 6,748, 6,191 and 4,652 images within each age category, respectively.

\begin{table}
\ra{1.2}
\begin{center}
\caption{Architecture of the generator.}
\label{tab:arch_generator}
\begin{tabular}{p{0.06\textwidth}|p{0.13\textwidth}|p{0.11\textwidth}|p{0.09\textwidth}}\toprule
\multicolumn{4}{c}{Encoder}                   \\ \hline
\hfil{\#Layer} & \hfil{Convolution}   & \hfil{Normalization}     & \hfil{Non-linear} \\ \hline
\hfil{1}       & \hfil{\textit{k=7, s=1, p=1}} & \hfil{Instance} & \hfil{ReLU}       \\ \hline
\hfil{2}       & \hfil{\textit{k=3, s=2, p=1}} & \hfil{Instance} & \hfil{ReLU}       \\ \hline 
\multicolumn{4}{c}{Residual Block ($\times$ 6)}        \\ \hline
\hfil{\#Layer} & \hfil{Convolution}   & \hfil{Normalization}     & \hfil{Non-linear} \\ \hline
\hfil{1}       & \hfil{\textit{k=3, s=2, p=1}} & \hfil{Instance} & \hfil{ReLU}       \\ \hline
\hfil{2}       & \hfil{\textit{k=3, s=2, p=1}} & \hfil{Instance} & \hfil{ReLU}       \\ \hline 
\multicolumn{4}{c}{Decoder}                   \\ \hline
\hfil{\#Layer} & \hfil{Deconvolution} & \hfil{Normalization}     & \hfil{Non-linear} \\ \hline
\hfil{1}       & \hfil{\textit{k=3, s=2, p=1}} & \hfil{Instance} & \hfil{ReLU}       \\ \hline
\hfil{2}       & \hfil{\textit{k=3, s=2, p=1}} & \hfil{Instance} & \hfil{Tanh}       \\ \bottomrule
\end{tabular}
\end{center}
\end{table}

\begin{table}
\ra{1.2}
\begin{center}
\caption{Architecture of the discriminators.}
\label{tab:arch_discriminator}
\begin{tabular}{p{0.06\textwidth}|p{0.13\textwidth}|p{0.11\textwidth}|p{0.09\textwidth}}\toprule
\multicolumn{4}{c}{Feature-Level ($\times$ 4)}        \\ \hline
\hfil{\#Layer} & \hfil{Fully-Connected}   & \hfil{Normalization}     & \hfil{Non-linear} \\ \hline
\hfil{1}       & \hfil{\textit{128}} & \hfil{Instance} & \hfil{LeakyReLU}       \\ \hline
\hfil{2}       & \hfil{\textit{64}} & \hfil{Instance} & \hfil{LeakyReLU}       \\ \hline 
\hfil{3}       & \hfil{\textit{32}} & \hfil{Instance} & \hfil{LeakyReLU}       \\ \hline 
\hfil{4}       & \hfil{\textit{16}} & \hfil{Instance} & \hfil{LeakyReLU}       \\ \hline 
\hfil{5}       & \hfil{\textit{1}} & \hfil{-} & \hfil{-}       \\ \hline
\multicolumn{4}{c}{Image-Level}                   \\ \hline
\hfil{\#Layer} & \hfil{Convolution}   & \hfil{Normalization}     & \hfil{Non-linear} \\ \hline
\hfil{1}       & \hfil{\textit{k=3, s=2, p=1}} & \hfil{Instance} & \hfil{LeakyReLU}       \\ \hline
\hfil{2}       & \hfil{\textit{k=3, s=2, p=1}} & \hfil{Instance} & \hfil{LeakyReLU}       \\ \hline 
\hfil{3}       & \hfil{\textit{k=3, s=2, p=1}} & \hfil{Instance} & \hfil{LeakyReLU}       \\ \hline 
\hfil{4}       & \hfil{\textit{k=3, s=2, p=1}} & \hfil{Instance} & \hfil{LeakyReLU}       \\ \hline 
\hfil{5}       & \hfil{\textit{k=3, s=1, p=1}} & \hfil{-} & \hfil{-}       \\ \bottomrule
\end{tabular}
\end{center}
\end{table}

%% ---------------------------------------------------------------------------- %%

\subsection{AOFS quality criteria}

There are two criteria commonly used to measure the quality of synthesized images \cite{yang2018learning,liu2019attribute} in the AOFS task. Under the first criterion, synthesized images are fed into an age category classifier to evaluate whether the depicted face has been transformed to the target age category. The second criterion measures the identity permanence and relies on face verification models to validate whether the synthesized image and the source image depict the same person. 

\begin{figure}[t]
\begin{center}
\includegraphics[width=1.\linewidth]{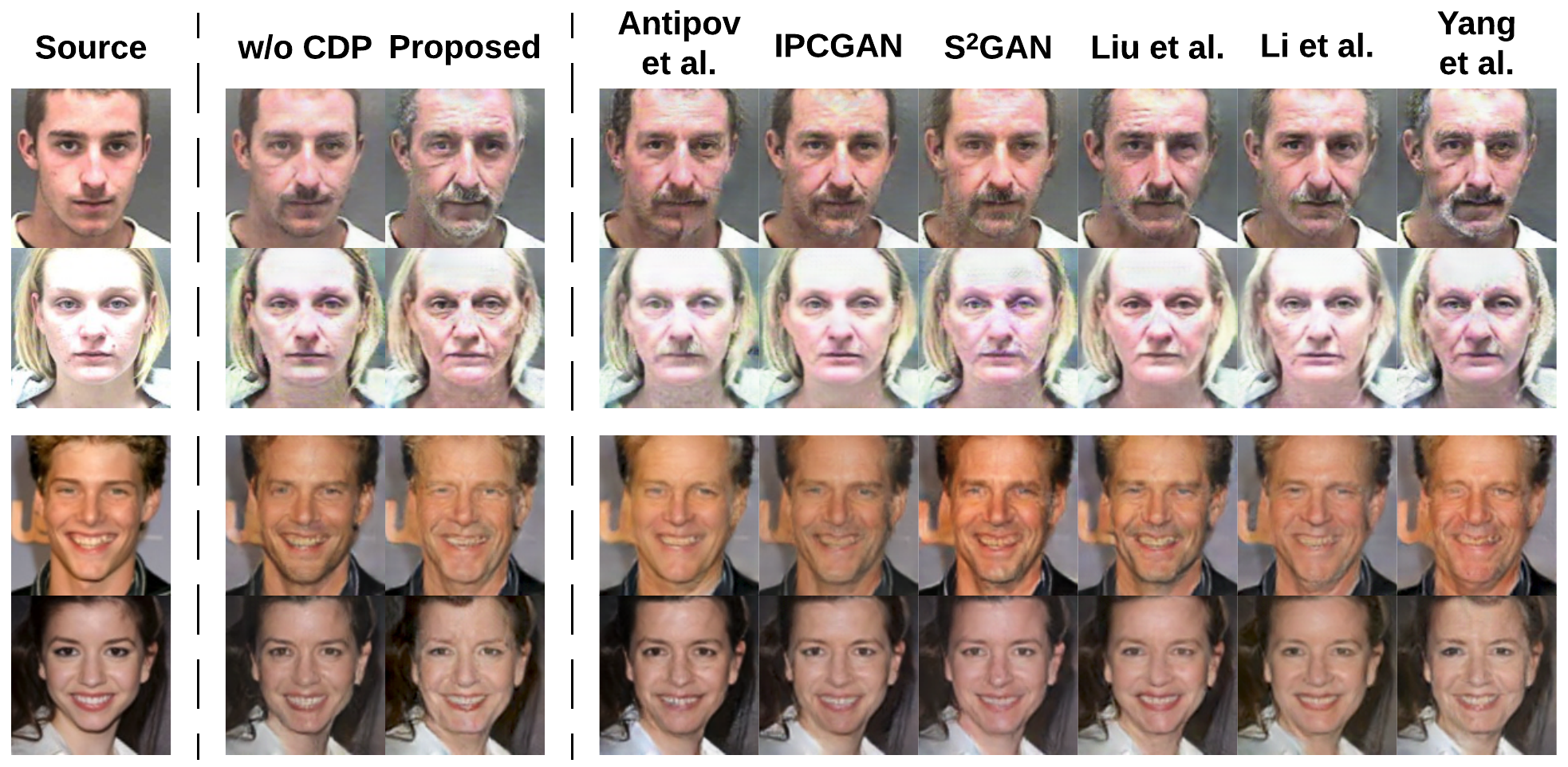}
\end{center}
   \caption{Visual comparison of a baseline model, six state-of-the-art works, and our proposed method on two benchmarks. The top two rows show the results on the MORPH II dataset and the bottom two rows show the results on the CACD. The input image is within the youngest group and the results are expected to be within the eldest group.}
\label{fig:baseline}
\end{figure}

To evaluate our method and demonstrate its robustness, we use another two large-scale benchmark datasets to train two separate validation networks, one for each criterion. In particular, we use the AgeDB dataset \cite{moschoglou2017agedb}, which is widely used for age estimation, to train the network that evaluates the synthesis accuracy and a face recognition benchmark dataset, the VGGFace2 dataset \cite{cao2018vggface2}, to train the network that evaluates the identity permanence capabilities. In addition, we use the commonly used ResNet-50 as the backbone for both evaluation networks.

\subsection{Network architecture}
% We adopt the architecture in \cite{wang2018orthogonal} for our MTFE with two branches of fully-connected layers, one for each task. 

We employ the architecture from~\cite{zhu2017unpaired} for our generator. The generator has six residual blocks and each convolutional and deconvolutional layer is followed by an instance normalization and a ReLU function. For the image-level discriminator, we implement a patch discriminator~\cite{isola2017image} with five convolutional layers, each followed by an instance normalization and a LeakyReLU function. Each feature-level discriminator has the same architecture as that of the image-level discriminator but consists of fully-connected layers. 

The details of the architectures of the generator and discriminators in our AOFS method are tabulated in Tables~\ref{tab:arch_generator} and \ref{tab:arch_discriminator}, respectively. In both tables, for each convolutional and deconvolutional layer, $k$ indicates the kernel size, $s$ indicates the stride, and $p$ indicates the padding size. In Table~\ref{tab:arch_discriminator}, the second column for the feature-level discriminators tabulates the dimensions of the corresponding layer.

\begin{table*}
\ra{1.2}
\caption{Age category classification accuracy (\%) on the images synthesized for the MORPH II dataset and the CACD for the aging process.}
\begin{center}
\begin{tabular}{p{0.15\textwidth}p{0.11\textwidth}p{0.11\textwidth}p{0.11\textwidth}p{0.005\textwidth}p{0.11\textwidth}p{0.11\textwidth}p{0.11\textwidth}} \toprule
%\multicolumn{9}{c}{Aging} \midrule
& \multicolumn{3}{c}{MORPH II} && \multicolumn{3}{c}{CACD}  \\
\cmidrule{2-4} \cmidrule{6-8}
%& \multicolumn{3}{c}{$w = 8$} & \phantom{abc}& \multicolumn{3}{c}{$w = 16$} &
%\phantom{abc} & \multicolumn{3}{c}{$w = 32$}\\
\hfil{Age Category} & \hfil{31-40} & \hfil{41-50} & \hfil{51$^+$} && \hfil{31-40} & \hfil{41-50} & \hfil{51$^+$} \\ \midrule
\hfil{Natural Faces} & \hfil{59.04 $\mypm$ 2.42} & \hfil{58.68 $\mypm$ 2.18} & \hfil{58.83 $\mypm$ 2.23} && \hfil{37.91 $\mypm$ 5.09} & \hfil{37.34 $\mypm$ 4.79} & \hfil{34.46 $\mypm$ 4.92} \\
\midrule
\hfil{Antipov \textit{et al.} \cite{antipov2017face}} & \hfil{39.56 $\mypm$ 2.28} & \hfil{39.79 $\mypm$ 2.10} & \hfil{35.22 $\mypm$ 2.50} && \hfil{20.29 $\mypm$ 4.58} & \hfil{20.49 $\mypm$ 5.04} & \hfil{18.43 $\mypm$ 5.40} \\
\hfil{IPCGAN~\cite{wang2018face}} & \hfil{44.67 $\mypm$ 2.25} & \hfil{44.70 $\mypm$ 2.43} & \hfil{41.84 $\mypm$ 1.77} && \hfil{24.90 $\mypm$ 4.29} & \hfil{27.70 $\mypm$ 4.25} & \hfil{28.49 $\mypm$ 5.00} \\
\hfil{S$^2$GAN \cite{he2019s2gan}} & \hfil{52.97 $\mypm$ 2.65} & \hfil{52.46 $\mypm$ 1.84} & \hfil{51.30 $\mypm$ 1.98} && \hfil{29.25 $\mypm$ 4.88} & \hfil{29.05 $\mypm$ 4.62} & \hfil{26.33 $\mypm$ 4.81} \\
\hfil{Liu \textit{et al.} \cite{liu2019attribute}} & \hfil{52.12 $\mypm$ 1.97} & \hfil{53.85 $\mypm$ 1.92} & \hfil{54.82 $\mypm$ 1.45} && \hfil{29.31 $\mypm$ 5.16} & \hfil{31.87 $\mypm$ 4.95} & \hfil{32.79 $\mypm$ 4.88} \\
\hfil{Li \textit{et al.} \cite{li2019global}} & \hfil{51.22 $\mypm$ 2.15} & \hfil{53.60 $\mypm$ 1.74} & \hfil{54.61 $\mypm$ 1.97} && \hfil{28.61 $\mypm$ 4.41} & \hfil{31.02 $\mypm$ 4.19} & \hfil{32.46 $\mypm$ 4.75} \\
\hfil{Yang \textit{et al.} \cite{yang2018learning}} & \hfil{53.24 $\mypm$ 1.67} & \hfil{53.23 $\mypm$ 2.86} & \hfil{53.20 $\mypm$ 1.73} && \hfil{30.68 $\mypm$ 4.12} & \hfil{30.85 $\mypm$ 4.43} & \hfil{31.64 $\mypm$ 4.38} \\
\hfil{w/o CDP} & \hfil{43.52 $\mypm$ 1.73} & \hfil{41.53 $\mypm$ 1.82} & \hfil{41.93 $\mypm$ 1.45} && \hfil{25.01 $\mypm$ 5.52} & \hfil{25.06 $\mypm$ 4.89} & \hfil{25.55 $\mypm$ 5.18} \\
\hfil{Proposed} & \textbf{\hfil{56.60 $\mypm$ 1.91}} & \textbf{\hfil{55.42 $\mypm$ 1.80}} & \textbf{\hfil{54.63 $\mypm$ 1.98}} && \textbf{\hfil{33.73 $\mypm$ 3.91}} & \textbf{\hfil{33.77 $\mypm$ 4.32}} & \textbf{\hfil{32.54 $\mypm$ 4.61}} \\
\bottomrule
\end{tabular}
\end{center}
\label{tab:aging}
\end{table*}

\subsection{Data augmentation}
When training the MTFE and validation networks, we use a combination of rotation, flip, and crop operations to augment the data. Specifically, we first randomly rotate each image by a angle between +10 deg. and -10 deg., and then randomly flip the rotated image with a probability of 0.5. Finally, we pad the image on all sides with 10 pixels and crop the padded image at a random location to the original image size (i.e. $128 \times 128$ pixels). When training the proposed AOFS method, in order to increase the size of the training set without introducing additional variance to the dataset, we only use the flip operation.

\subsection{Hyper-parameter setting}
When training the MTFE, we set the batch size to 128 and the initial learning rate to $0.002$ for both datasets. We train it for 500 epochs while decreasing the learning rate by 0.1 every 150 epochs. When training the AOFS method, we set the batch size to 8 and the initial learning rate to $0.0002$. The learning rate decreases linearly after the first 25 epochs. We empirically set $\lambda_{adv_{feature}}$ to $1$ and $\lambda_{AT}$ to $0.001$. The margin hyper-parameter, $m$ in Eq. (\ref{adversarial-triplet_AOFS}), is set to 0.3. We use the PyTorch framework \cite{paszke2017automatic} for the implementation and run each experiment for 50 epochs. All experiments are run on a single NVIDIA GTX2080Ti GPU. 

% When training the feature extractor, we set the batch size to 128 and an initial learning rate of $0.002$ for both datasets. When training the age-oriented face synthesis model, we set the batch size to 8 and the initial learning rate to $0.0002$. For both datasets, we empirically set $\lambda_{adv_{feature}}$ to $1$, $\lambda_{DP-Triplet}$ to $0.001$ and $\lambda_{dp}$ to $1$. Two margin hyper-parameters, $m1$ and $m2$, are both set to 0.3. We use PyTorch framework \cite{paszke2017automatic} for the implementation and run 50 epochs for each experiment. 

\subsection{Synthesis accuracy}
We first qualitatively evaluate the synthesized facial images based on their visual quality. We then present quantitative results based on age category classification accuracy, image quality and the degree of mode collapse. We perform these evaluations for our AOFS method and several state-of-the-art methods.

\begin{table*}
\ra{1.2}
\caption{Age category classification accuracy (\%) on the images synthesized for the MORPH II dataset and the CACD for the rejuvenating process.}
\begin{center}
\begin{tabular}{p{0.15\textwidth}p{0.11\textwidth}p{0.11\textwidth}p{0.11\textwidth}p{0.005\textwidth}p{0.11\textwidth}p{0.11\textwidth}p{0.11\textwidth}} \toprule
%\multicolumn{9}{c}{Aging} \midrule
& \multicolumn{3}{c}{MORPH II} && \multicolumn{3}{c}{CACD}  \\
\cmidrule{2-4} \cmidrule{6-8}
%& \multicolumn{3}{c}{$w = 8$} & \phantom{abc}& \multicolumn{3}{c}{$w = 16$} &
%\phantom{abc} & \multicolumn{3}{c}{$w = 32$}\\
\hfil{Age Category} & \hfil{30$^-$} & \hfil{31-40} & \hfil{41-50} && \hfil{30$^-$} & \hfil{31-40} & \hfil{41-50} \\ \midrule
\hfil{Natural Faces} & \hfil{63.08 $\mypm$ 1.81} & \hfil{59.04 $\mypm$ 2.42} & \hfil{58.68 $\mypm$ 2.18} && \hfil{43.82 $\mypm$ 4.06} & \hfil{37.91 $\mypm$ 5.09} & \hfil{37.34 $\mypm$ 4.79} \\
\midrule
\hfil{Antipov \textit{et al.} \cite{antipov2017face}} & \hfil{50.55 $\mypm$ 2.32} & \hfil{44.71 $\mypm$ 2.45} & \hfil{44.77 $\mypm$ 1.84} && \hfil{28.41 $\mypm$ 3.92} & \hfil{26.36 $\mypm$ 5.87} & \hfil{26.17 $\mypm$ 4.71} \\
\hfil{IPCGAN~\cite{wang2018face}} & \hfil{57.33 $\mypm$ 1.82} & \hfil{52.03 $\mypm$ 1.79} & \hfil{52.32 $\mypm$ 2.21} && \hfil{32.67 $\mypm$ 4.43} & \hfil{31.89 $\mypm$ 4.50} & \hfil{31.41 $\mypm$ 5.08} \\
\hfil{S$^2$GAN \cite{he2019s2gan}} & \hfil{58.18 $\mypm$ 1.83} & \hfil{54.11 $\mypm$ 2.04} & \hfil{54.24 $\mypm$ 1.43} && \hfil{33.36 $\mypm$ 4.01} & \hfil{32.30 $\mypm$ 4.38} & \hfil{32.63 $\mypm$ 3.89} \\
\hfil{Liu \textit{et al.} \cite{liu2019attribute}} & \hfil{59.06 $\mypm$ 2.41} & \hfil{55.33 $\mypm$ 1.61} & \hfil{55.54 $\mypm$ 2.01} && \hfil{36.65 $\mypm$ 4.31} & \hfil{34.25 $\mypm$ 4.34} & \hfil{34.26 $\mypm$ 4.69} \\
\hfil{Li \textit{et al.} \cite{li2019global}} & \hfil{58.87 $\mypm$ 2.30} & \hfil{55.21 $\mypm$ 2.18} & \hfil{55.06 $\mypm$ 1.94} && \hfil{37.84 $\mypm$ 4.66} & \hfil{34.95 $\mypm$ 4.86} & \hfil{34.30 $\mypm$ 4.26} \\
\hfil{Yang \textit{et al.} \cite{yang2018learning}} & \hfil{60.79 $\mypm$ 2.21} & \hfil{56.99 $\mypm$ 2.17} & \textbf{\hfil{56.65 $\mypm$ 2.39}} && \hfil{39.09 $\mypm$ 4.72} & \hfil{35.62 $\mypm$ 4.83} & \hfil{35.89 $\mypm$ 4.61} \\
\hfil{w/o CDP} & \hfil{53.67 $\mypm$ 2.35} & \hfil{51.41 $\mypm$ 2.33} & \hfil{51.96 $\mypm$ 2.45} && \hfil{29.17 $\mypm$ 5.05} & \hfil{28.42 $\mypm$ 5.39} & \hfil{28.67 $\mypm$ 5.31} \\
\hfil{Proposed} & \textbf{\hfil{61.20 $\mypm$ 1.41}} & \textbf{\hfil{57.12 $\mypm$ 1.36}} & \hfil{56.55 $\mypm$ 2.23} && \textbf{\hfil{41.24 $\mypm$ 4.12}} & \textbf{\hfil{36.84 $\mypm$ 4.10}} & \textbf{\hfil{36.59 $\mypm$ 4.81}} \\
\bottomrule
\end{tabular}
\end{center}
\label{tab:rejuvenating}
\end{table*}

\begin{table}
\ra{1.20}
\begin{center}
\caption{ResNet Score and Fr\'echet ResNet Distance on the MORPH II dataset.}
\label{tab:score_morph}
\begin{tabular}{p{0.15\textwidth}p{0.11\textwidth}p{0.11\textwidth}}\toprule
\hfil{Model} & \hfil{RS} & \hfil{FRD} \\ \midrule
\hfil{Antipov \textit{et al.} \cite{antipov2017face}} & \hfil{27.83 $\mypm$ 1.34} & \hfil{31.72  $\mypm$ 0.60} \\
\hfil{IPCGAN~\cite{wang2018face}} & \hfil{36.70 $\mypm$ 1.18} & \hfil{28.08 $\mypm$ 0.44} \\
\hfil{S$^2$GAN \cite{he2019s2gan}} & \hfil{38.92 $\mypm$ 1.14} & \hfil{25.64 $\mypm$ 0.32} \\
\hfil{Liu \textit{et al.} \cite{liu2019attribute}} & \hfil{39.14 $\mypm$ 1.23} & \hfil{25.57 $\mypm$ 0.42} \\
\hfil{Li \textit{et al.} \cite{li2019global}} & \hfil{39.26 $\mypm$ 1.22} & \hfil{25.51 $\mypm$ 0.41} \\
\hfil{Yang \textit{et al.} \cite{yang2018learning}} & \hfil{43.35 $\mypm$ 1.36} & \hfil{22.30 $\mypm$ 0.59} \\
\hfil{w/o CDP} & \hfil{30.19 $\mypm$ 1.26} & \hfil{28.62 $\mypm$ 0.49} \\
\hfil{Proposed} & \textbf{\hfil{44.04 $\mypm$ 1.25}} & \textbf{\hfil{21.93 $\mypm$ 0.46}} \\
\bottomrule
\end{tabular}
\end{center}
\end{table}

\begin{table}
\ra{1.20}
\begin{center}
\caption{ResNet Score and Fr\'echet ResNet Distance on the CACD.}
\label{tab:score_cacd}
\begin{tabular}{p{0.15\textwidth}p{0.11\textwidth}p{0.11\textwidth}}\toprule
\hfil{Model} & \hfil{RS} & \hfil{FRD} \\ \midrule
\hfil{Antipov \textit{et al.} \cite{antipov2017face}} & \hfil{24.71 $\mypm$ 2.04} & \hfil{33.83  $\mypm$ 0.95} \\
\hfil{IPCGAN~\cite{wang2018face}} & \hfil{33.21 $\mypm$ 1.82} & \hfil{30.18 $\mypm$ 0.79} \\
\hfil{S$^2$GAN \cite{he2019s2gan}} & \hfil{34.24 $\mypm$ 1.75} & \hfil{27.01 $\mypm$ 0.61} \\
\hfil{Liu \textit{et al.} \cite{liu2019attribute}} & \hfil{34.54 $\mypm$ 1.86} & \hfil{26.99 $\mypm$ 0.63} \\
\hfil{Li \textit{et al.} \cite{li2019global}} & \hfil{35.00 $\mypm$ 1.91} & \hfil{26.91 $\mypm$ 0.67} \\
\hfil{Yang \textit{et al.} \cite{yang2018learning}} & \hfil{37.39 $\mypm$ 2.09} & \hfil{24.62 $\mypm$ 0.87} \\
\hfil{w/o CDP} & \hfil{30.87 $\mypm$ 1.87} & \hfil{30.71 $\mypm$ 0.82} \\
\hfil{Proposed} & \textbf{\hfil{38.55 $\mypm$ 1.90}} & \textbf{\hfil{23.98 $\mypm$ 0.73}} \\
\bottomrule
\end{tabular}
\end{center}
\end{table}

\begin{table}
\ra{1.20}
\begin{center}
\caption{Degree of mode collapse as measured by the KL divergence.}
\label{tab:kl}
\begin{tabular}{p{0.15\textwidth}p{0.11\textwidth}p{0.11\textwidth}}\toprule
\hfil{Model} & \hfil{MORPH II} & \hfil{CACD} \\ \midrule
\hfil{Antipov \textit{et al.} \cite{antipov2017face}} & \hfil{1.86 $\mypm$ 0.10} & \hfil{1.93  $\mypm$ 0.13} \\
\hfil{IPCGAN~\cite{wang2018face}} & \hfil{0.64 $\mypm$ 0.15} & \hfil{0.68 $\mypm$ 0.21} \\
\hfil{S$^2$GAN \cite{he2019s2gan}} & \hfil{0.59 $\mypm$ 0.08} & \hfil{0.62 $\mypm$ 0.11} \\
\hfil{Liu \textit{et al.} \cite{liu2019attribute}} & \hfil{0.55 $\mypm$ 0.09} & \hfil{0.57 $\mypm$ 0.13} \\
\hfil{Li \textit{et al.} \cite{li2019global}} & \hfil{0.55 $\mypm$ 0.11} & \hfil{0.58 $\mypm$ 0.14} \\
\hfil{Yang \textit{et al.} \cite{yang2018learning}} & \hfil{0.49 $\mypm$ 0.04} & \hfil{0.52 $\mypm$ 0.05} \\
\hfil{w/o CDP} & \hfil{1.19 $\mypm$ 0.09} & \hfil{1.30 $\mypm$ 0.14} \\
\hfil{Proposed} & \textbf{\hfil{0.37 $\mypm$ 0.04}} & \textbf{\hfil{0.42 $\mypm$ 0.07}} \\
\bottomrule
\end{tabular}
\end{center}
\end{table}

\subsubsection{Visual Quality}

Fig. \ref{fig:progress_aging} and \ref{fig:progress_rejuvenating} show some sample images synthesized by our AOFS method. Fig. \ref{fig:progress_aging} shows aging results for 6 subjects from the MORPH II dataset and 6 from the CACD using a source image from the youngest category ($30^-$). We can see that our method turns hair gray or white, introduces forehead wrinkles and nasolabial folds, and makes the skin to appear rough. Fig. \ref{fig:progress_rejuvenating} shows rejuvenating results for 6 subjects from each dataset using a source image from the oldest category ($51^+$). We can see that for these cases, our method removes wrinkles and gray/white hair.

We also evaluate six state-of-the-art methods, namely the method by Antipov \textit{et al.} \cite{antipov2017face}, the Identity-Preserving Conditional Generative Adversarial Networks (IPCGAN) \cite{wang2018face}, the S$^2$GAN \cite{he2019s2gan}, and the methods by Liu \textit{et al.} \cite{liu2019attribute}, Li \textit{et al.} \cite{li2019global}, and Yang \textit{et al.} \cite{yang2018learning}. To have a fair comparison, we replace the feature extractors in these methods with our pre-trained MTFE and use the same number of residual blocks in their generator expect for the method in \cite{antipov2017face}, as there is no residual block originally involved in this particular method. 

% The results reported in this work for these methods are obtained by using the official implementations.

Since the synthesis accuracy of our AOFS method depends on the CDP, we also evaluate a baseline model without the CDP (hereinafter called \textit{w/o CDP}) as part of an ablation study. The \textit{w/o CDP} model replaces the CDP with a simple feature-level discriminator, which makes this model similar to a vanilla GAN but with two discriminators, one at the feature level and the other at the image level. 

% Moreover, by using our Conditionally Selective Discriminator mechanism, the model can alleviate the mode collapse issue with plentiful synthesised ageing effects.

Fig. \ref{fig:baseline} depicts the visual results of these evaluations. Note that it is visually evident that the results generated by the \textit{w/o CDP} model do not contain much aging and rejuvenating effects as this model suffers from the mode collapse issue. On the contrary, our proposed method can synthesize the aging and rejuvenating effects realistically. Among all state-of-the-art methods, Yang \textit{et al.} \cite{yang2018learning} is able to synthesize the most realistic effects due to the use of a multi-level feature discriminator.

\begin{table*}
\ra{1.20}
\caption{Face verification results in terms of accuracy (\%) for the MORPH II dataset and the CACD. The query images are the original facial images, and the gallery images are the synthesised images generated by each corresponding model.}
\begin{center}
\begin{tabular}{p{0.05\textwidth}p{0.12\textwidth}p{0.1\textwidth}p{0.1\textwidth}p{0.1\textwidth}p{0.001\textwidth}p{0.1\textwidth}p{0.1\textwidth}p{0.1\textwidth}} \toprule
%\multicolumn{9}{c}{Ageing} \midrule
&& \multicolumn{3}{c}{Aging} && \multicolumn{3}{c}{Rejuvenating}  \\
\cmidrule{3-5} \cmidrule{7-9}
%& \multicolumn{3}{c}{$w = 8$} & \phantom{abc}& \multicolumn{3}{c}{$w = 16$} &
%\phantom{abc} & \multicolumn{3}{c}{$w = 32$}\\
\multicolumn{2}{c}{Gallery Image} & \hfil{S31-40} & \hfil{S41-50} & \hfil{S51$^+$} && \hfil{S41-50} & \hfil{S31-40} & \hfil{S30$^-$} \\ \midrule
\multicolumn{1}{c:}{\multirow{9}{*}{MORPH II}} & \hfil{Antipov \textit{et al.} \cite{antipov2017face}} & \hfil{94.46 $\mypm$ 0.16} & \hfil{93.57 $\mypm$ 0.12} & \hfil{91.24 $\mypm$ 0.20} && \hfil{95.33 $\mypm$ 0.16} & \hfil{93.54 $\mypm$ 0.13} & \hfil{92.48 $\mypm$ 0.27} \\
\multicolumn{1}{c:}{}&\hfil{IPCGAN~\cite{wang2018face}} & \hfil{94.56 $\mypm$ 0.23} & \hfil{93.87 $\mypm$ 0.19} & \hfil{91.63 $\mypm$ 0.22} && \hfil{94.91 $\mypm$ 0.28} & \hfil{93.83 $\mypm$ 0.20} & \hfil{92.21 $\mypm$ 0.27} \\
\multicolumn{1}{c:}{}&\hfil{S$^2$GAN \cite{he2019s2gan}} & \hfil{94.88 $\mypm$ 0.09} & \hfil{93.65 $\mypm$ 0.17} & \hfil{91.44 $\mypm$ 0.12} && \hfil{95.50 $\mypm$ 0.11} & \hfil{94.72 $\mypm$ 0.19} & \hfil{92.54 $\mypm$ 0.18} \\
\multicolumn{1}{c:}{}&\hfil{Liu \textit{et al.} \cite{liu2019attribute}} & \hfil{94.22 $\mypm$ 0.28} & \hfil{93.49 $\mypm$ 0.26} & \hfil{91.28 $\mypm$ 0.21} && \hfil{95.63 $\mypm$ 0.22} & \hfil{94.84 $\mypm$ 0.23} & \hfil{93.23 $\mypm$ 0.27} \\
\multicolumn{1}{c:}{}&\hfil{Li \textit{et al.} \cite{li2019global}} & \hfil{95.08 $\mypm$ 0.11} & \hfil{93.99 $\mypm$ 0.14} & \hfil{91.87 $\mypm$ 0.15} && \hfil{95.40 $\mypm$ 0.14} & \hfil{94.05 $\mypm$ 0.16} & \hfil{92.52 $\mypm$ 0.17} \\
\multicolumn{1}{c:}{}&\hfil{Yang \textit{et al.} \cite{yang2018learning}} & \hfil{94.29 $\mypm$ 0.22} & \hfil{93.34 $\mypm$ 0.27} & \hfil{91.18 $\mypm$ 0.28} && \hfil{95.76 $\mypm$ 0.21} & \hfil{94.40 $\mypm$ 0.22} & \hfil{93.76 $\mypm$ 0.29} \\
\multicolumn{1}{c:}{}&\hfil{Triplet} & \hfil{97.87 $\mypm$ 0.07} & \hfil{97.01 $\mypm$ 0.09} & \hfil{94.86 $\mypm$ 0.17} && \hfil{98.14 $\mypm$ 0.06} & \hfil{98.23 $\mypm$ 0.11} & \hfil{97.71 $\mypm$ 0.14} \\
\multicolumn{1}{c:}{}&\hfil{Proposed} & \textbf{\hfil{99.06 $\mypm$ 0.03}} & \textbf{\hfil{98.73 $\mypm$ 0.06}} & \textbf{\hfil{95.58 $\mypm$ 0.11}} && \textbf{\hfil{99.61 $\mypm$ 0.03}} & \textbf{\hfil{99.39 $\mypm$ 0.08}} & \textbf{\hfil{97.85 $\mypm$ 0.09}} \\
\midrule
\multicolumn{1}{c:}{\multirow{9}{*}{CACD}} & \hfil{Antipov \textit{et al.} \cite{antipov2017face}} & \hfil{92.06 $\mypm$ 0.27} & \hfil{88.46 $\mypm$ 0.35} & \hfil{85.40 $\mypm$ 0.56} && \hfil{92.67 $\mypm$ 0.23} & \hfil{89.30 $\mypm$ 0.28} & \hfil{86.24 $\mypm$ 0.42} \\
\multicolumn{1}{c:}{}&\hfil{IPCGAN~\cite{wang2018face}} & \hfil{92.29 $\mypm$ 0.30} & \hfil{88.77 $\mypm$ 0.33} & \hfil{85.22 $\mypm$ 0.57} && \hfil{93.93 $\mypm$ 0.25} & \hfil{89.32 $\mypm$ 0.32} & \hfil{85.35 $\mypm$ 0.50} \\
\multicolumn{1}{c:}{}&\hfil{S$^2$GAN \cite{he2019s2gan}} & \hfil{92.39 $\mypm$ 0.35} & \hfil{88.94 $\mypm$ 0.55} & \hfil{85.87 $\mypm$ 0.59} && \hfil{93.32 $\mypm$ 0.33} & \hfil{89.60 $\mypm$ 0.42} & \hfil{86.29 $\mypm$ 0.54} \\
\multicolumn{1}{c:}{}&\hfil{Liu \textit{et al.} \cite{liu2019attribute}} & \hfil{92.25 $\mypm$ 0.26} & \hfil{88.51 $\mypm$ 0.32} & \hfil{85.46 $\mypm$ 0.48} && \hfil{93.21 $\mypm$ 0.23} & \hfil{89.50 $\mypm$ 0.32} & \hfil{85.02 $\mypm$ 0.47} \\
\multicolumn{1}{c:}{}&\hfil{Li \textit{et al.} \cite{li2019global}} & \hfil{93.33 $\mypm$ 0.24} & \hfil{89.04 $\mypm$ 0.38} & \hfil{85.91 $\mypm$ 0.45} && \hfil{94.52 $\mypm$ 0.21} & \hfil{89.47 $\mypm$ 0.36} & \hfil{85.31 $\mypm$ 0.39} \\
\multicolumn{1}{c:}{}&\hfil{Yang \textit{et al.} \cite{yang2018learning}} & \hfil{92.24 $\mypm$ 0.29} & \hfil{88.58 $\mypm$ 0.48} & \hfil{85.54 $\mypm$ 0.57} && \hfil{92.80 $\mypm$ 0.20} & \hfil{89.07 $\mypm$ 0.39} & \hfil{86.91 $\mypm$ 0.42} \\
\multicolumn{1}{c:}{}&\hfil{Triplet} & \hfil{93.89 $\mypm$ 0.17} & \hfil{92.73 $\mypm$ 0.21} & \hfil{89.15 $\mypm$ 0.24} && \hfil{94.79 $\mypm$ 0.15} & \hfil{93.46 $\mypm$ 0.17} & \hfil{90.31 $\mypm$ 0.23} \\
\multicolumn{1}{c:}{}&\hfil{Proposed} & \textbf{\hfil{94.98 $\mypm$ 0.10}} & \textbf{\hfil{94.16 $\mypm$ 0.14}} & \textbf{\hfil{90.77 $\mypm$ 0.18}} && \textbf{\hfil{95.08 $\mypm$ 0.11}} & \textbf{\hfil{94.56 $\mypm$ 0.14}} & \textbf{\hfil{91.68 $\mypm$ 0.15}} \\
\bottomrule
\end{tabular}
\end{center}
\label{tab:identity}
\end{table*}

\subsubsection{Age category classification accuracy}

Table~\ref{tab:aging} and \ref{tab:rejuvenating} tabulate the age category classification accuracies of various methods on the synthesized images when images from the $30^-$ and $51^+$ categories are used as source images, respectively. In these tables, the \textit{Natural Faces} row tabulates the accuracy attained when using the original facial images. Since \cite{antipov2017face} uses a relatively shallow generator compared to other works, its performance is hence below others by a significant margin. IPCGAN uses the age labels as conditions in the GAN learning process and incorporates an age category classification loss. However, due to the fact that the classification error is high (the classifier is noisy), the gradient for the age information is not accurate. As a result, although its performance is higher than that of \cite{antipov2017face}, it is still lower than the one attained on the original facial images by a large margin. The recently proposed S$^2$GAN attains a higher accuracy by implementing a customized generator where each age category is associated with a decoder. The methods of Liu \textit{et al.} \cite{liu2019attribute} and Li \textit{et al.} \cite{li2019global} achieve similar accuracy since both use the Wavelet transform. Among all the other evaluated methods, the one proposed by Yang \textit{et al.} \cite{yang2018learning} achieves the best performance by using a multi-level feature discriminator. By adding a feature-level discriminator to the vanilla GAN, the baseline \textit{w/o CDP} model achieves a comparable performance to that achieved by IPCGAN. Our proposed AOFS method outperforms all evaluated methods for the majority of age categories.

\subsubsection{Image Quality}

The synthesis accuracy is also related to the quality of the generated images \cite{wang2018face}. The quality and diversity of the synthesized images are usually measured in terms of the Inception Score (IS) and the Fr\'echet Inception Distance (FID). IS measures the image quality and diversity by computing the KL divergence between the real and the generated class distributions. On the other hand, FID uses a multivariate Gaussian distribution to model the data distribution and the mean and the covariance from two distributions to compute their distance. Since we use a ResNet-50 to evaluate the identity permanence capabilities (see Section IV.G), we rename these two metrics as the ResNet Score (RS) and the Fr\'echet ResNet Distance (FRD). The RS and FRD are tabulated in Table~\ref{tab:score_morph} and Table~\ref{tab:score_cacd}, respectively, for our AOFS method and several state-of-the-art methods. Since our AOFS method can render more realistic aging and rejuvenating effects than other evaluated methods and has stronger identity permanence capabilities, it achieves the best performance for both metrics, especially for the FRD, which is sensitive to the mode collapse issue.

\subsubsection{Degree of Mode Collapse}

Since our method tackles the AOFS task from the aspect of mode learning, we also measure the degree of mode collapse by computing the KL divergence between the distribution of the synthesized images and the expected distribution. We compute this divergence for all synthesized images within each fold.

As shown in Table~\ref{tab:kl}, the proposed AOFS method significantly outperforms the baseline model and the method in \cite{antipov2017face}, which use the negative log-likelihood loss from the vanilla GAN. By using different discriminators to learn different modes, our method also achieves a lower divergence value compared to other methods that leverage the least square loss from the LSGAN.

% To further demonstrate the effectiveness of the Conditionally Selective Discriminator mechanism, we use the t-SNE to visualise the distributions of images from the original dataset and synthesised images on the feature space. In Figure~\ref{fig:distribution}, we use four colours to indicate images within four age categories. It is clear that images synthesised by \textit{w/o FD} are gathered around one location, i.e. mode. By adopting one feature-level discriminator, the distribution of synthesised images moves towards the real distribution. In addition, by implementing our proposed Conditionally Selective Discriminator mechanism, the synthesised images have a similar distribution compared to the real distribution.

% \begin{figure*}
% \begin{center}
% \includegraphics[width=17.5cm]{distribution.pdf}
% \end{center}
%   \caption{Data distribution on the MORPH II dataset. From left to right: images from the original dataset; synthesised images from \textit{w/o FD}; synthesised images from \textit{w/o Pool}; synthesised images from proposed model.}
% \label{fig:distribution}
% \end{figure*}

\subsection{Identity permanence}
% In this part, we first present a toy example to show the distribution learned from our proposed DP-Triplet loss. Then, we show the face verification accuracy on the synthesised images.

% \begin{figure}
% \begin{center}
% \includegraphics[width=1\linewidth]{mnist_train.pdf}
% \end{center}
%   \caption{Feature distributions on the MNIST training set. Left: features learned by Triplet loss; Right: features learned by DP-Triplet loss.}
% \label{fig:id_train}
% \end{figure}

% \begin{figure}[t]
% \begin{center}
% \includegraphics[width=1\linewidth]{mnist_test.pdf}
% \end{center}
%   \caption{Feature distributions on the MNIST test set. Left: features learned by Triplet loss; Right: features learned by DP-Triplet loss.}
% \label{fig:id_test}
% \end{figure}

To evaluate the identity permanence on the synthesized images, we design a new baseline, the \textit{Triplet} model. Specifically, in the \textit{Triplet} model, we replace the Adversarial Triplet loss with the original Triplet loss to directly compare these two loss functions. The identity permanence capabilities are measured in terms of the face verification accuracy, i.e.. whether the synthesized image and the original image depict the same person. To this end, we define three input settings based on three different target age categories for each synthesis process. Specifically, the query images are the original facial images from the datasets, while the gallery images are the synthesized images that are expected to be within the target age category, as tabulated in Table \ref{tab:identity} with the column headings \textit{S31-40}, \textit{S41-50}, and \textit{S51$^+$} for the aging process and headings \textit{S41-50}, \textit{S31-40}, and \textit{S30$^-$} for the rejuvenating process. For example, \textit{S31-40} refers to the synthesized images expected to be within the $31-40$ category. We use the \textit{cosine similarity} to measure the distance of each pair of query and gallery images.

As tabulated in Table \ref{tab:identity}, all the state-of-the-art methods achieve a similar accuracy since they all use a similar strategy, namely, minimizing the distance between two identity-specific features using the L1 or L2 loss. Li \textit{et al.} \cite{li2019global} slightly outperforms other methods as it uses a combination of these two losses. The subtle difference in accuracy among these methods may also be due to the quality of the images, since the identity information may be distorted in images of poor quality. By replacing the L1 or L2 loss with the Triplet loss, the identity permanence capability can be remarkably boosted by about 3 \% on both datasets. Our AOFS method, which uses the Adversarial Triplet loss, reduces intra-class variances within each age category in the feature space. Consequently, it achieves the highest accuracy among all evaluated methods.

\section{Conclusion}
In this paper, we tackle the Age-Oriented Face Synthesis task from the aspect of the mode learning. Specifically, we present an AOFS method that incorporates a novel Conditional Discriminator Pool to alleviate the mode collapse issue in the vanilla GAN. Our method also incorporates a novel Adversarial Triplet loss to attain strong identity permanence capabilities. By using the proposed CDP, only the target feature-level discriminator that learns the current mode is deployed, which does not increase the computational complexity during training. Our CDP then allows learning multiple modes explicitly and independently. As a result, our proposed AOFS method outperforms several state-of-the-art methods on AOFS benchmark datasets. In the future, we will investigate into improving the aging and rejuvenating effects by including the synthesis and removal of wrinkles and face shape manipulation among different age categories. Improving these aspects of the synthesis process is expected to further boost the synthesis accuracy and have the potential to simulate a more personalized aging and rejuvenating process. 

\section{Acknowledgment}
This work is supported by the EU Horizon 2020 - Marie Sklodowska-Curie Actions through the project Computer Vision
Enabled Multimedia Forensics and People Identification (Project No. 690907, Acronym: IDENTITY).

\ifCLASSOPTIONcaptionsoff
  \newpage
\fi



%\bibliographystyle{IEEEtran}

%-------------------------------------------------------------------------
%%%%%%%%% References
\bibliographystyle{IEEEbib}
\bibliography{refs}

\end{document}